\documentclass{article}
\usepackage{PRIMEarxiv}

\usepackage[utf8]{inputenc} 
\usepackage[T1]{fontenc}    
\usepackage{hyperref}       
\usepackage{url}            
\usepackage{booktabs}       
\usepackage{amsfonts}       
\usepackage{nicefrac}       
\usepackage{microtype}      
\usepackage{lipsum}
\usepackage{fancyhdr}       
\usepackage{graphicx}       
\graphicspath{{media/}}     
 \usepackage{natbib}

\usepackage{amsmath}
\usepackage{amsthm}  
\usepackage{multirow}

\usepackage{pifont}
\usepackage{newunicodechar}
\usepackage{tabularx}
\usepackage{float}
\usepackage{longtable}
\usepackage{subcaption}

\newunicodechar{✓}{\ding{51}}
\newunicodechar{✗}{\ding{55}}

\usepackage{chngcntr}
\counterwithin{theorem}{section}
\counterwithin{proposition}{section}
\counterwithin{lemma}{section}
\counterwithin{corollary}{section}

\usepackage{graphicx}
\usepackage[font=small,labelfont=bf]{caption}
\usepackage{subcaption} 
\graphicspath{{fig/}}   

\usepackage{booktabs,tabularx}
\usepackage[ruled,vlined,linesnumbered]{algorithm2e}
\SetKwInput{KwIn}{Input}
\SetKwInput{KwOut}{Output}
\SetKw{KwRet}{return}
\SetKw{KwBreak}{break}
\DontPrintSemicolon

\title{Training Memory in Deep Neural Networks: Mechanisms, Evidence, and Measurement Gaps

}

\author{
  Vasileios Sevetlidis* \\
  Athena RC \\
  Xanthi, GR67100, Greece\\
  \texttt{vasiseve@athenarc.gr} \\
   \AND
  George Pavlidis \\
  Athena RC \\
  University Campus Kimmeria \\
  Xanthi, GR67100, Greece\\
  \texttt{gpavlid@athenarc.gr} \\
}

\begin{document}
\maketitle

\begin{abstract}
Modern deep-learning training is not memoryless. Updates depend on optimizer moments and averaging, data-order policies (random reshuffling vs with-replacement, staged augmentations and replay), the nonconvex path, and auxiliary state (teacher EMA/SWA, contrastive queues, BatchNorm statistics). This survey organizes mechanisms by source, lifetime, and visibility. It introduces seed-paired, function-space causal estimands; portable perturbation primitives (carry/reset of momentum/Adam/EMA/BN, order-window swaps, queue/teacher tweaks); and a reporting checklist with audit artifacts (order hashes, buffer/BN checksums, RNG contracts). The conclusion is a protocol for portable, causal, uncertainty-aware measurement that attributes how much training history matters across models, data, and regimes.

\end{abstract}


\section{Introduction}

Training a deep neural network is \emph{not} memoryless. By \textit{training memory} we mean that what the learner does next depends not only on its current parameters and the current minibatch, but also on \emph{how it arrived there}—the recent sequence of updates and data. Specially, training memory can arise from (i) \emph{optimizer state} (e.g., momentum buffers or adaptive moments that carry summaries of past gradients), (ii) \emph{sampler state} (e.g., the order in which examples are presented or priorities that make some examples more likely to be seen), and (iii) the \emph{parameter path} itself (the route through the loss landscape) which, in nonconvex problems, makes the order of small updates matter.

Classical and modern optimization analyses formalize these ideas. Momentum and averaging explicitly carry information forward across steps, and in nonconvex settings stochastic updates do \emph{not} commute—applying update $A$ then $B$ can lead to a different point than applying $B$ then $A$—so outcomes can depend on path and ordering \citep{bottou2018optimization,sutskever2013importance,polyak1992averaging,mishchenko2020random,gurbuzbalaban2021rr}. In practice, stateful mechanisms are used deliberately at scale: adaptive optimizers (e.g., Adam/AdamW), exponential moving averages of weights (EMA) and stochastic weight averaging (SWA), and sharpness-aware updates introduce controlled history dependence to stabilize or improve generalization \citep{izmailov2018swa,foret2020sharpness}. In distributed and federated settings, server-side adaptivity further exposes the role of carried-over state across rounds \citep{reddi2021fedopt}.

Memory also lives in the data pipeline. Even without explicit priorities, \emph{random reshuffling} (processing each example once per epoch in a random order) behaves differently from with-replacement sampling; theory and experiments show distinct convergence properties and, at times, performance \citep{shamir2016without,gurbuzbalaban2021rr,mishchenko2020random}. When sampling is \emph{stateful}—for example via online or prioritized selection that revisits high-loss or high-influence examples—training can speed up, but attribution becomes entangled between data order and optimizer history \citep{loshchilov2015online,katharopoulos2018importance,johnson2018training,Schaul2016PER}. Beyond loss curves, studies of representation similarity (e.g., SVCCA and CKA) document that learned features can \emph{drift} across runs or phases, even when top-line metrics look similar \citep{raghu2017svcca,kornblith2019cka,ding2021grounding}. These tools describe \emph{that} change occurs, but they are not causal diagnostics: they do not attribute changes to optimizer state, sampler state, or path effects, and they are rarely paired with effect sizes and uncertainty.

Despite broad recognition that DNN training has memory, the community lacks a \emph{portable}, \emph{causal}, and \emph{standardized} way to quantify \emph{how much} that memory matters across architectures, data sets, and training regimes. Typical reports emphasize final accuracy or loss and often leave protocols under-specified (e.g., whether momentum was reset between phases, the exact data-order policy, or augmentation schedules). Order- and state-driven phenomena then become difficult to reproduce or compare. Related areas reinforce this picture at longer time scales: in continual learning, forgetting and transfer depend on task order and replay \citep{parisi2019continual,delange2021survey}; in curriculum learning, pacing and example ordering affect stability and sample efficiency \citep{soviany2022curriculum}. Yet across these settings there is still no consensus diagnostic that cleanly attributes effects to optimizer state, sampler memory, or parameter-path dependence.

We synthesize mechanisms and evidence for training memory across optimizers, samplers, and parameter paths, and we surface protocol pitfalls that hinder attribution. This article does not propose a new algorithm or a single diagnostic. Instead, it contributes (i) a taxonomy (source–lifetime–visibility), (ii) a synthesis of theory and evidence, (iii) causal estimands with portable perturbation primitives, and (iv) a reporting checklist that make attribution auditable. We then articulate solution-agnostic \emph{desiderata} for future diagnostics: isolate sources via controlled perturbations; report effect sizes in function space alongside standard metrics; track representation drift with appropriate caveats; and emphasize early indicators that predict late generalization. Our aim is to motivate principled, causal measurement of training memory—without committing to a specific methodology in this study.

\section{Relation to Prior Surveys}
Several survey families intersect with what we term \emph{training memory}—persistent traces of optimization history, algorithmic state, and data ordering that shape the final solution. Geometry-oriented overviews on loss landscapes and model merging illuminate path dependence and mergeability (e.g., linear/curvilinear connectivity and permutation-aware alignment), but typically abstract away the concrete sources of \emph{state} accumulated during training \citep{ferbach2024lmc,li2023deepfusion,yang2024modelmerging}. Work on sampler order shows that random reshuffling (without replacement) can materially alter convergence and final solutions relative to with-replacement SGD, yet these analyses are not usually framed as a unifying “memory” mechanism nor tied to broader reporting practice \citep{mishchenko2020random,safran2020sgd,yu2023rr}. Reviews of normalization emphasize that running statistics and batch coupling are consequential sources of hidden state—with recent evidence of task-specific pitfalls—again treated largely in isolation from other mechanisms \citep{huang2023normalization,rivoir2024bnpitfalls}. In self-supervised and contrastive learning, surveys document explicit memory structures such as queues and banks (e.g., MoCo-style dictionaries), but scope is confined to SSL rather than integrated with optimizer/geometry/order effects \citep{uelwer2025ssl}. Complementary lines synthesize representational and functional similarity tools for auditing training trajectories \citep{klabunde2025similarity}, ensembling surveys that touch SWA/SWAG as temporal averaging along a path \citep{yang2023ensemble}, federated learning surveys that catalog server/client momentum as cross-round state \citep{chen2024federatedcsur}, and calibration/uncertainty surveys that focus on reliability outcomes rather than the upstream memory mechanisms that might drive them \citep{wang2023calibration}. Documentation artifacts such as Model Cards, Datasheets, and NeurIPS reproducibility checklists advance general reporting, but they do not yet target memory-sensitive details (e.g., sampler semantics, EMA/SWA configuration, BN/SyncBN regimes, queue refresh rules) \citep{mitchell2019modelcards,gebru2021datasheets,pineau2021reproducibilityreport}.

Our survey differs by (i) unifying these strands under a single taxonomy of \emph{training memory} that treats optimizer momentum/EMA/SWA, sampler order/reshuffling, normalization running statistics, explicit buffers (queues/replay), geometric path dependence, and federated server/client state as first-class, interacting carriers of history; (ii) mapping cross-mechanism interactions (e.g., how order shapes EMA trajectories or how BN state mediates mergeability); and (iii) proposing \emph{measurement protocols} and \emph{reporting checklists} specific to memory. Concretely, we standardize trajectory-level similarity profiling, interpolation/merging and barrier tests, order-sensitivity ablations, state ablations for EMA/SWA/BN/queues, and calibration under controlled memory manipulations, alongside a minimal set of fields for declaring and stress-testing memory effects. This integrated perspective complements existing surveys by making the memory mechanisms explicit, comparable, and reportable across training pipelines.

\section{Taxonomy of Training Memory}
\label{sec:taxonomy}

\noindent
A learner has \emph{training memory} when its update at step $t$ depends on more than the current weights and minibatch—it also depends on how we got here (optimizer state, data-order decisions, or the specific path taken through parameter space). Formally, relative to the visible interface $(\theta_t,b_t)$ i.e., the state the outer training loop exposes: current weights $\theta_t$ and current minibatch $b_t$, we say the procedure has memory at time $t$ if the distribution of the next update depends on history beyond that interface:
\[
\mathbb{P}\!\left(\Delta\theta_t \,\middle|\, \theta_t,\, b_t,\, \mathcal{H}_t\right)\;\neq\; \mathbb{P}\!\left(\Delta\theta_t \,\middle|\, \theta_t,\, b_t\right),
\]
where $\theta_t$ are the current weights, $b_t$ is the current minibatch, and $\mathcal{H}_t$ aggregates prior randomness, batches, and internal states up to time $t$. This reconciles the intuition with a Markov view: if we augment the state $S$ to
\[
S_t \;=\; \big(\theta_t,\; \text{optimizer buffers}_t,\; \text{sampler RNG/order}_t,\; \text{external queues/banks}_t,\ldots\big),
\]
then $\mathbb{P}(\Delta\theta_t \mid S_t)$ can be history-independent by construction. “Having memory’’ thus means effects persist when those augmented components are not controlled or are only partially observed.

\begin{figure}[t]
  \centering
  \includegraphics[width=\linewidth]{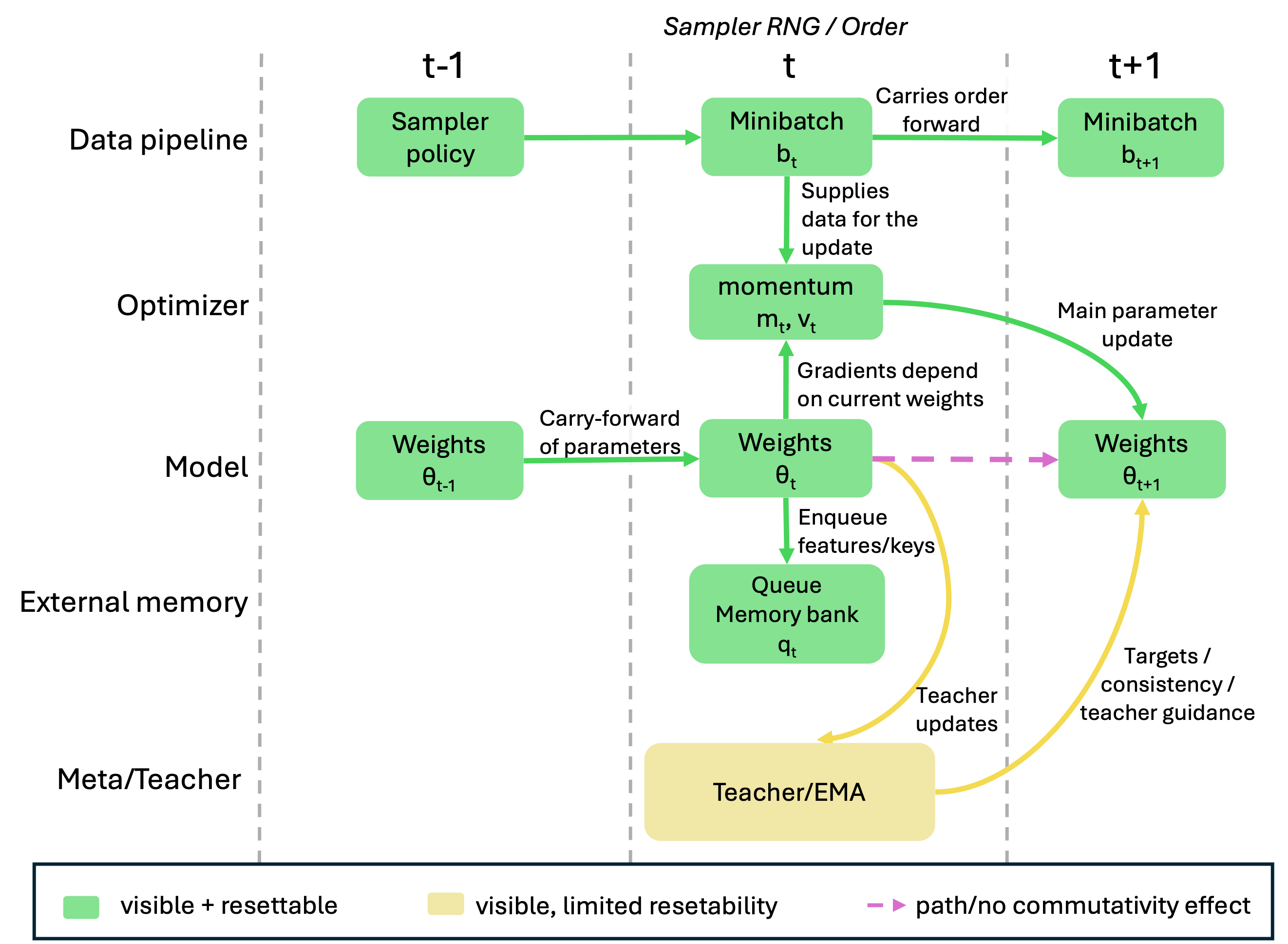}
  \caption{Schematic of the training loop with explicit and implicit state. Boxes: weights $\theta_t$, optimizer buffers (e.g., momentum/Adam), sampler RNG/order, external queues/memory banks, and teacher/EMA. Arrows indicate what carries across steps and phases; coloring encodes visibility/resetability.}
  \label{fig:train-loop-states}
\end{figure}

To orient newcomers, we organize the space along three axes: \textbf{source} (where the memory comes from), \textbf{lifetime} (how long it persists), and \textbf{visibility} (can we see or reset it). \figureautorefname\ref{fig:train-loop-states} illustrates the explicit and implicit states we consider in the training loop.

\subsection{Axis 1 — Source (where memory comes from)}
\noindent
\textbf{(S1) Optimizer / trajectory state.}  
Modern optimizers carry running summaries of the past. Momentum accumulates a smoothed direction of travel; Adam/AdamW keep first/second moments; EMA/SWA average weights along a trajectory; sharpness-aware methods adjust steps using recent curvature; second-order/preconditioned methods keep structured curvature statistics. Each mechanism makes the \emph{next} step a function of \emph{many previous} steps.  If we reset or perturb the optimizer’s internal state (e.g., zero momentum buffers; clear EMA), downstream behavior can shift even when weights and data are unchanged.

\quad\textit{Examples.}  
\emph{Momentum/Nesterov} keep an exponential moving average (EMA) of gradients; the decay $\beta$ induces a practical “half-life” (how many steps a gradient still matters) \citep{sutskever2013importance,polyak1992averaging}.  
\emph{Adam/AdamW} maintain biased/unbiased moment estimates; AdamW decouples weight decay from the loss gradient, explicitly injecting history via moments while controlling norm growth \citep{kingma2017adam,loshchilov2018adamw}.  
\emph{Averaging} (Polyak averaging; SWA) smooths iterate noise and tends to land in wider basins \citep{polyak1992averaging,izmailov2018swa}.  
\emph{SAM} alters steps to avoid sharp regions, making updates depend on recent local geometry \citep{foret2020sharpness}.  
\emph{Limited-memory second-order} (e.g., K-FAC, Shampoo) maintain layer- or tensor-structured curvature state that persists across steps \citep{martens2015optimizing,gupta2018shampoo}.

\noindent
\textbf{(S2) Sampler / data-order state.}  The data pipeline is not memoryless. Changing \emph{which} examples appear (and \emph{when}) changes the gradient noise and the path. Random reshuffling (each example once per epoch) behaves differently from with-replacement sampling; curricula/pacing alter difficulty over time; prioritized sampling and replay make exposure frequencies stateful.  Holding optimizer fixed, changes in order, pacing, or priority can move training toward different regions of the landscape.

BatchNorm keeps running means/variances that persist across updates and are used at evaluation \citep{ioffe2015batchnorm}. Re-estimating these statistics after a change in weights or domain measurably shifts performance: SWA explicitly requires recomputing BN stats for the averaged model \citep{izmailov2018swa,maddox2019swag}; domain-adaptation methods replace source BN statistics with target-domain estimates (AdaBN), improving accuracy \citep{li2016adabn}; and test-time adaptation often updates BN statistics on the fly \citep{wang2021tent}. Thus BN stats are explicit, resettable state that should be logged/manipulated alongside momentum/Adam buffers. By contrast, LayerNorm normalizes per example and uses no running averages, providing a useful foil \citep{ba2016layernorm}.
Many widely used augmentation policies are explicitly time-varying and thus act like \emph{stateful sampling}: because the transformation distribution changes across epochs, the effective minibatch distribution drifts even when the dataset is fixed. AutoAugment learns stage-specific policies, while RandAugment provides magnitude and count parameters that are often scheduled during training. In attribution studies, these schedules should be treated on the same footing as order policies—frozen or perturbed in isolation—since they alter which \emph{views} of examples are seen when \citep{cubuk2019autoaugment,cubuk2020randaugment}.

\quad\textit{Examples.}  
\emph{Random reshuffling vs.\ with-replacement} have distinct convergence behaviors and often different empirical performance \citep{shamir2016without,gurbuzbalaban2021rr,mishchenko2020random}.  
\emph{Curricula/pacing} implement long-horizon ordering (easy$\rightarrow$hard; staged augmentation) \citep{soviany2022curriculum}.  
\emph{Prioritized/importance sampling} upsamples “informative” examples (high loss/gradient norm, TD-error), trading speed for entanglement between sampler and optimizer state \citep{katharopoulos2018importance,johnson2018training,Schaul2016PER}.  
\emph{Replay/coresets} retain subsets over time, making the sampler explicitly stateful across many steps or tasks \citep{parisi2019continual}.

\noindent
\textbf{(S3) Parameter-path dependence.}  
 In nonconvex landscapes, small updates need not commute: taking step $A$ then $B$ can end somewhere else than $B$ then $A$. Even with a “stateless” optimizer and IID minibatches, the \emph{route} conditions the \emph{destination}. This shows up as preference for flatter minima or distinct but mode-connected solutions.  
 If two runs start from the same initialization but follow different orderings or schedules, they can converge to functionally different solutions.

A practical way to \emph{make path effects visible} is via mode-connectivity diagnostics: fit linear or low-curvature paths between solutions found under different orders/schedules and probe predictions along the path. When there are low-loss connectors, we can test whether the behavior of the function space varies smoothly or exhibits 'kinks' that reveal qualitatively different solutions; lack of connectivity suggests truly distinct basins \citep{garipov2018loss,draxler2018essentially,frankle2020lmc}. In short, connectivity provides an operational visibility tool for (S3): it does not only assert that the path matters; it lets us \emph{measure} how endpoints relate in the landscape.

\quad\textit{Examples.}  
\emph{Loss landscape geometry} and \emph{mode connectivity} document low-loss paths between solutions and relate flatter regions to generalization \citep{li2018visualizing,garipov2018loss}.  
\emph{Solution anchoring} in transfer: L2-SP pulls fine-tuning towards the pre-train weights; EWC constrains movement along important directions - both bake history into the objective \citep{li2018l2sp,kirkpatrick2017overcoming}.

\noindent
\textbf{(S4) Architectural / external memory.}  
 Some training recipes add \emph{extra state} beyond model weights: feature queues, memory banks, or plastic traces. Updates then depend on this side memory as well as the current batch.  
 Clearing the queue or bank changes gradients immediately; longer queues imply longer memory. 

In contrastive pipelines with queues or memory banks, the queue length sets an explicit lifetime: with a FIFO dictionary of size $K$ and minibatch size $B$, a stored key typically persists for $\approx K/B$ updates before eviction; ablations in momentum-contrast systems show that varying $K$ materially changes both optimization and downstream performance \citep{he2020moco,wu2018instancedisc}.

\quad\textit{Examples.}  
\emph{Contrastive queues/memory banks} (MoCo; instance discrimination) maintain evolving representations outside the main weights \citep{he2020moco,wu2018instancedisc}.  
\emph{Hebbian/plastic traces} introduce additional, training-updated state that shapes learning dynamics \citep{halvagal2023hebbian}.

\noindent
\textbf{(S5) Meta-state (teachers / learned optimizers).}  
 A teacher network (often an EMA of the student) or a learned optimizer evolves over time and steers updates across many steps.  
Resetting or slowing the teacher/outer-loop often changes stability and final performance.

\quad\textit{Examples.}  
\emph{Teacher–student EMA} (“Mean Teacher”) provides slowly moving targets that encode long-range history \citep{tarvainen2017meanteacher}.  
\emph{Learned/slow-fast optimizers} (learned optimizers; Lookahead) accumulate cross-step state and modulate inner updates \citep{andrychowicz2016l2l,zhang2019lookahead}.  
\emph{Federated training} exposes server/client accumulators as explicit cross-round memory \citep{mcmahan2017fedavg,reddi2021fedopt}. Concrete, source-specific perturbations that isolate (S1)–(S5)
are listed in \tableautorefname~\ref{tab:perturbation-primitives} (see §\ref{subsec:primitives}).
\subsection{Axis 2 — Lifetime (how long memory persists)}
\label{subsec:lifetime}

\begin{figure*}[t]
  \centering
  \includegraphics[width=\textwidth]{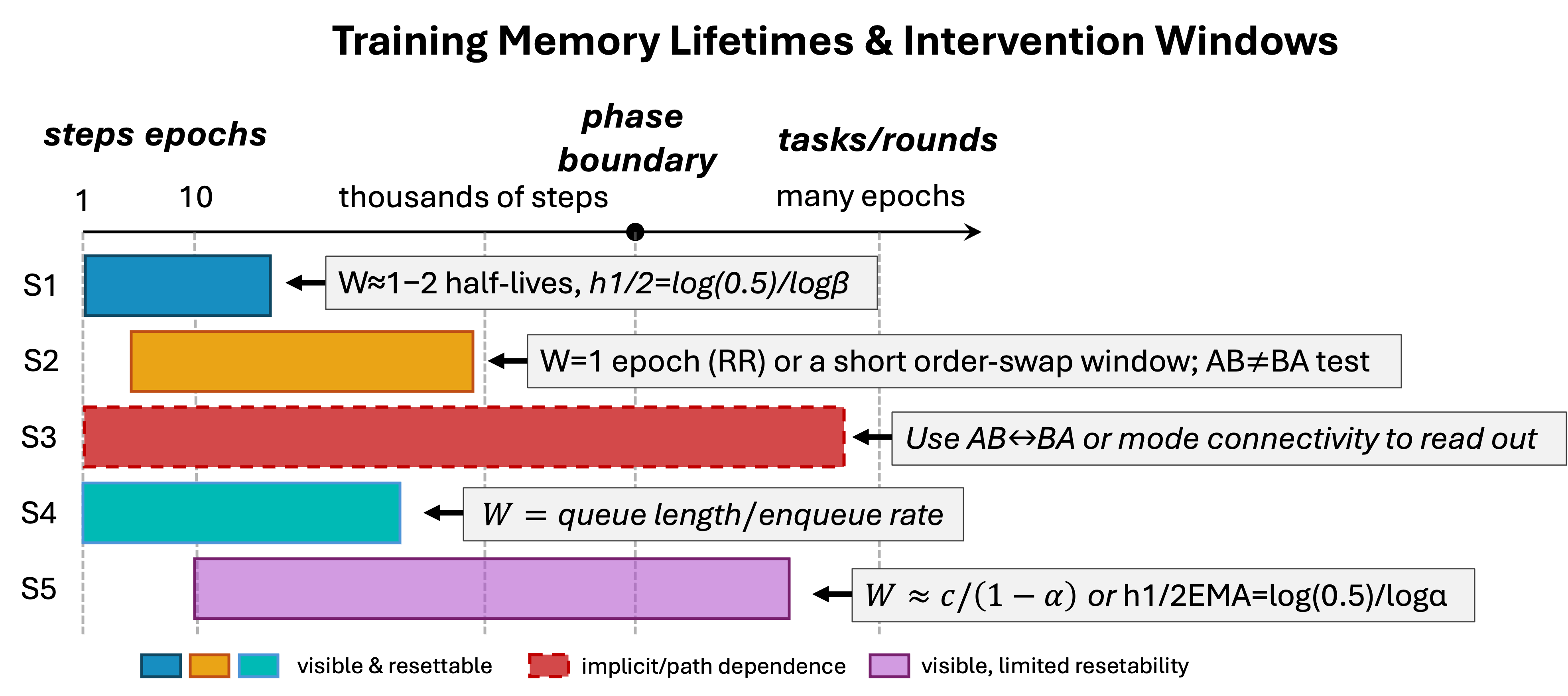}%
  \caption{Typical memory lifetimes by source (S1–S5), from steps $\rightarrow$ epochs $\rightarrow$ phases $\rightarrow$ tasks/rounds.
  Each band shows a suggested intervention window $W$ suited to that source:
  step-scale state (e.g., momentum/Adam, EMA) use $W\!\approx\!1$–$2$ half-lives; epoch-scale order use one full reshuffled epoch or a fixed minibatch window under with-replacement sampling; phase-scale effects probe the first $k$ epochs post-boundary; external queues use the queue \emph{turnover} (queue length divided by enqueue rate); short AB$\neq$BA order swaps at boundaries expose non-commutativity.}
  \label{fig:timescales}
\end{figure*}

As summarized in \figureautorefname~\ref{fig:timescales}, not all forms of training memory last the same amount of time.
Some mechanisms fade within a few dozen updates; others persist across an epoch, a phase boundary (pretraining$\!\to\!$fine-tuning), or even entire task sequences. Thinking in terms of \emph{lifetimes} explains why two runs that look identical locally can diverge globally.

\paragraph{Step scale.}
Optimizer statistics that decay every update create short-horizon memory. Momentum and the first/second moments in Adam/AdamW retain influence over tens to hundreds of steps depending on the decay; their practical half-life is $h_{1/2}=\log(0.5)/\log\beta$ (see \tableautorefname~\ref{tab:half-life}) \citep{sutskever2013importance,kingma2017adam,loshchilov2018adamw}. Weight averaging (EMA/SWA) integrates a trailing tail of iterates and thus also encodes recent history \citep{polyak1992averaging,izmailov2018swa,maddox2019swag}. These mechanisms make warm restarts behave differently from cold starts and explain why clearing buffers or freezing EMA for a short window can measurably alter early dynamics and calibration.

\paragraph{Epoch scale.}
The data pipeline dominates at the epoch horizon. With the dataset fixed, \emph{order} (random reshuffling vs.\ with-replacement), batching, and staged augmentations change the gradient noise and steer the path through parameter space \citep{shamir2016without,gurbuzbalaban2021rr,mishchenko2020random}. Curricula/pacing and replay policies induce persistence across many updates because the composition of minibatches co-varies over time \citep{soviany2022curriculum,Rebuffi2017iCaRL,LopezPaz2017GEM}. These effects often survive small hyperparameter tweaks precisely because their lifetime is longer than step-scale statistics.

\paragraph{Phase scale.}
Explicit program boundaries—e.g., pretraining$\!\to\!$fine-tuning, or schedule restarts—carry history across many thousands of steps. The initialization checkpoint anchors fine-tuning; penalties such as L2-SP/EWC make this anchoring explicit \citep{li2018l2sp,kirkpatrick2017overcoming}. Meta-state like teacher EMA/SWA can further strengthen it if carried across the boundary \citep{tarvainen2017meanteacher,izmailov2018swa}. Comparisons across fine-tunes should therefore report whether optimizer buffers and EMA/SWA were \emph{carried} or \emph{reset}, and whether BatchNorm statistics were recalibrated \citep{ioffe2015batchnorm,izmailov2018swa,maddox2019swag,li2016adabn,wang2021tent}. To expose phase-scale memory, branch at the boundary and apply the intervention for the first $k$ epochs of the new phase (typically $k\in[1,5]$).

\paragraph{Task/round scale.}
In continual and federated settings, memory spans tasks or rounds. Replay buffers and consolidation penalties retain information across tasks \citep{parisi2019continual,delange2021survey}, while client sampling and server-side accumulators (momentum/Adam variants) retain cross-round state \citep{mcmahan2017fedavg,reddi2021fedopt,Karimireddy2020Scaffold,Li2020FedProx}. Here, outcomes depend on the \emph{sequence} of tasks/clients as well as the optimizer trajectory.

\paragraph{Choosing $W$.}
In practice, $W$ is selected with reference to the characteristic lifetime of the perturbed source. For step-scale statistics, a practical choice is $W \approx 1$–$2$ half-lives. For order interventions, one reshuffled epoch under random-reshuffling (RR), or a fixed window under with-replacement sampling, is often sufficient. Following a phase boundary, it can be informative to confine the intervention to the first $k$ epochs. For external-memory mechanisms, taking $W$ commensurate with the queue turnover (queue length divided by enqueue rate) reflects the horizon over which contents refresh \citep{he2020moco,wu2018instancedisc}. Short AB$\neq$BA swaps at natural boundaries may illuminate non-commutativity while limiting interference with the overall training program \citep{shamir2016without,mishchenko2020random,gurbuzbalaban2021rr}.

\subsection{Axis 3 — Visibility (can we see or reset it?)}
\label{subsec:visibility}

\begin{figure}[t]
  \centering
  \includegraphics[width=0.63\linewidth]{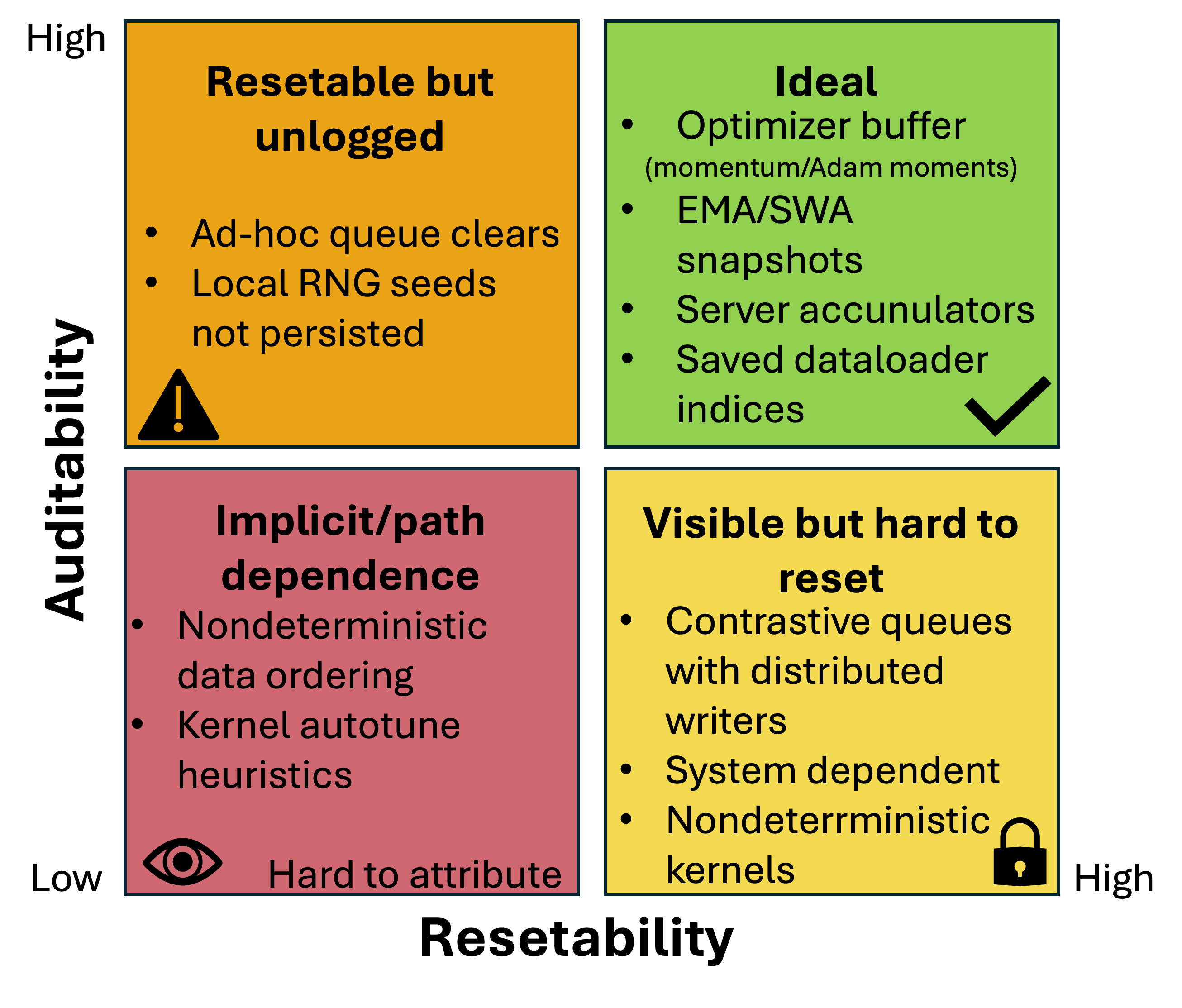}
  \caption{Resetability $\times$ auditability matrix with examples: ideal (visible, resetable, auditable) vs.\ visible-but-hard-to-reset, resetable-but-unlogged, and implicit/path dependence.}
  \label{fig:reset-audit}
\end{figure}

Some forms of memory are \emph{visible} as concrete state we can inspect or zero (e.g., momentum/Adam buffers, EMA/SWA snapshots, contrastive queues), while others are only detectable through behavior and geometry (path dependence in nonconvex landscapes). Two practical dials matter for protocol design: \emph{resetability} (can we deterministically clear or reinitialize the state in this framework?) and \emph{auditability} (is the state \emph{logged} or checksummed so others can verify what was carried across phases?). Resetability is imperfect in common stacks due to nondeterministic kernels and ops \footnote{Reproducibility — PyTorch Documentation: \url{https://pytorch.org/docs/stable/notes/randomness.html}, Accessed 2025-09-21}, so protocols should state what was reset and how it was verified. Auditability is helped by reproducibility checklists and reporting artifacts (e.g., order hashes, buffer-state logs, model/dataset documentation) advocated in community reports and documentation frameworks \citep{pineau2021reproducibilityreport,mitchell2019modelcards,gebru2021datasheets}. \figureautorefname\ref{fig:reset-audit} positions common examples on the resetability–auditability plane.

\emph{Explicit} memory includes any state materialized in the training loop. Momentum and Adam buffers, EMA/SWA snapshots, feature queues or memory banks used by contrastive methods, and server accumulators in federated training are all examples \citep{kingma2017adam,izmailov2018swa,he2020moco,mcmahan2017fedavg}. Because these states are tangible, they create clear levers for experimentation—one can log, reset, or ablate them—and clear responsibilities for reporting: whether such state is carried across epochs or phase boundaries often determines reproducibility.

\emph{Implicit} memory leaves no named buffer to read off. In overparameterized, nonconvex models, small stochastic updates need not commute; taking step $A$ then $B$ can land in a different region than $B$ then $A$. The history is thus written into the \emph{path} itself: which basin was entered, how flat the surrounding region is, and whether distinct solutions are mode-connected by low-loss corridors \citep{li2018visualizing,garipov2018loss}. When explicit state is controlled yet outcomes still diverge, the culprit is typically this path dependence—an interaction between geometry and ordering that is invisible unless one probes function space or representations, not just losses.

Two protocol–critical dimensions make the visibility axis actionable.
\emph{Resetability} asks whether a state can be deterministically zeroed or rewound (e.g., momentum/Adam buffers; EMA/SWA snapshots; dataloader order via saved index streams). Frameworks expose determinism toggles and seeded RNG streams, but nondeterministic kernels and backend heuristics still require explicit handling for faithful resets \citep{henderson2018matters}. 
\emph{Auditability} asks whether the state is logged so downstream analysis and reproduction are possible: per-epoch order hashes, buffer norms/checkpoints, and versioned configuration artifacts. Community guidance (NeurIPS Reproducibility Checklist; model cards; datasheets) emphasizes that audit logs are as important as code for post-hoc attribution \citep{pineau2021reproducibilityreport,mitchell2019modelcards,gebru2021datasheets}. 
Taken together, resetability (can we intervene?) and auditability (can we verify?) turn visibility into protocol design: visible but not resetable (hard to attribute); resetable but not audited (hard to trust); visible, resetable, and auditable (ideal).

Most practices blend the aforementioned. A run may use sharpness-aware updates and random reshuffling while maintaining a contrastive queue. The optimizer and sampler expose explicit dials; the landscape supplies the implicit backdrop on which those dials act. Any measurement or diagnostic that claims to attribute “training memory’’ needs to respect this composition: accounting for visible state while recognizing that part of what persists is the route the model took to get where it is. Any diagnostic that \emph{attributes} training memory should therefore: (i) separate optimizer vs.\ sampler vs.\ path effects; (ii) respect their timescales; and (iii) report effect sizes with uncertainty, not just point estimates.

\section{Theory \& Established Results}
\label{sec:theory}

This section explains \emph{why} training memory should exist in modern learning pipelines and \emph{what} the main theoretical lenses say about it. We focus on families of results that are widely cited and useful for reasoning about history-dependence, without committing to any particular diagnostic.

\subsection{Stochastic approximation and momentum: why recent history matters}

Classical stochastic approximation (SA) views training as a noisy recursion that tracks a moving target under sampling noise \citep{robbins1951sa,bottou2018optimization}. Once we introduce exponential smoothing, the recursion becomes explicitly history-dependent. In momentum SGD, a velocity $v_t=\beta v_{t-1}+(1-\beta)g_t$ accumulates gradients $g_t$ with geometric weights; Adam extends this idea with first and second moments \citep{sutskever2013importance,kingma2017adam}. The influence of a past gradient decays like $\beta^{\,h}$ after $h$ steps, so the \emph{effective half-life} of memory is
\[
h_{1/2} \;=\; \frac{\log 0.5}{\log \beta}\,.
\]
To ground this quantity, \tableautorefname~\ref{tab:half-life} translates common $\beta$ values into half-lives. We will use these numbers to set the short intervention window $W$ for optimizer-state perturbations (typically $W\!\approx\!1$–$2$ half-lives; see §\ref{subsec:primitives}). For example, $\beta{=}0.99$ implies $h_{1/2}\!\approx\!69$ steps, so a temporary momentum reset or EMA freeze for $W\!\in\![70,140]$ steps exposes the intended memory without long-term interference. 

Polyak--Ruppert averaging makes a similar point at the \emph{iterate} level: averaging a trailing tail of parameters reduces variance and nudges the solution toward wider, flatter regions \citep{polyak1992averaging}. Modern weight averaging (SWA) can be seen as a practical incarnation of this principle \citep{izmailov2018swa}. Taken together, SA and averaging formalize an obvious practitioner lesson: even if we freeze data and hyperparameters, changing the optimizer’s internal state (buffers, EMAs) changes what happens next because we have changed the \emph{recent past} the algorithm is carrying.

\begin{table}[t]
\centering
\small
\renewcommand{\arraystretch}{1.05}
\begin{tabular}{lrrrr}
\toprule
Momentum $\beta$ & 0.90 & 0.95 & 0.99 & 0.999 \\
\midrule
Half-life $h_{1/2}=\log(0.5)/\log(\beta)$ (steps) & 6.58 & 13.53 & 68.97 & 692.82 \\
\bottomrule
\end{tabular}
\caption{Optimizer memory half-life as a function of $\beta$. We use these values to pick the perturbation window $W$ for optimizer-state interventions ($W\!\approx\!1$–$2$ half-lives; see §\ref{subsec:primitives}).}
\label{tab:half-life}
\end{table}

A complementary view treats SGD as a stochastic sampler whose noise level steers \emph{path selection}. In the small–step limit, SGD can be approximated by an SDE whose effective “temperature’’ grows with the learning rate and shrinks with batch size, biasing trajectories toward wider basins; with explicit noise (SGLD), this interpretation becomes literal \citep{mandt2017sgdSDE,smith2018bayesian,welling2011sgld}. Practically, this ties optimizer and sampler parameter—step size, batch size, and even order-induced noise—to where the run goes in parameter space, motivating our emphasis on \emph{function-space} readouts (not just loss) when comparing memoryful training policies.

Lower noise (e.g., very large batches at fixed step size) can bias training toward \emph{sharper} minima, whereas higher noise explores and can escape them—consistent with reports that large-batch training often finds sharper solutions with worse test performance \citep{keskar2017largebatch}. This sampler-as-thermostat view dovetails with SGLD and Bayesian perspectives that tie batch size and step size to an effective temperature and predictive uncertainty \citep{welling2011sgld,smith2018bayesian}.

\subsection{Order dependence and non-commutativity: why the route matters}
In nonconvex objectives, small stochastic updates generally do \emph{not} commute. Applying minibatch $A$ and then $B$ yields a different point than $B$ then $A$ because the curvature encountered in between is different. Theory detects this even in convex baselines through the comparison of \emph{with-replacement} sampling and \emph{random reshuffling} (RR, without replacement): the two procedures induce different noise structures and can have different convergence rates, with RR often enjoying tighter guarantees \citep{shamir2016without,gurbuzbalaban2021rr,mishchenko2020random}. More recent analyses treat single-shuffle and arbitrary-order schedules, reinforcing that ``data order’’ is a real algorithmic choice, not an implementation detail \citep{koloskova2023arbitrary}. For our purposes, these results justify treating the sampler (and its ordering across epochs) as a bona fide \emph{source of memory}: by fixing or perturbing order, we change the path—and in nonconvex problems, changing the path changes the endpoint.

\subsection{Importance sampling and prioritization: memory in the sampler}
If examples are sampled with nonuniform probabilities $p_i$, unbiased risk estimates require reweighting by $1/p_i$. This can reduce gradient variance and accelerate optimization, particularly when losses or gradient norms are heavy-tailed \citep{katharopoulos2018importance,csiba2018minibatchis}. At scale, robust or approximate prioritization schemes trade exactness for speed \citep{johnson2018training}. Beyond convergence speed, however, deep models exhibit more nuanced behavior: importance weighting interacts with loss curvature and with adaptive optimizers, and its effect on \emph{generalization} depends on when and how it is applied (e.g., early vs.\ late training, separable vs.\ nonseparable regimes) \citep{byrd2019importanceweighting}. The conceptual takeaway is simple: once the sampler’s probabilities depend on the evolving state (losses, errors, features), the sampler acquires \emph{its own memory}. Which examples are repeatedly emphasized becomes part of the history that shapes the model.

\subsection{Continual learning and the stability--plasticity trade-off: intended long-term memory}
The stability--plasticity dilemma asks the learner to acquire new information (plasticity) without erasing useful prior knowledge (stability) \citep{grossberg2013art}. Modern continual-learning methods make long-term memory explicit via three broad strategies: (i) regularization that keeps solutions near previous optima (e.g., EWC), (ii) replay buffers or coresets that preserve a working memory of past data, and (iii) architectural expansion that isolates parameters for new tasks \citep{kirkpatrick2017overcoming,parisi2019continual,delange2021survey}. These mechanisms are intentionally stateful: their purpose is to carry information across task boundaries. They therefore provide concrete testbeds for studying training memory at the longest timescales.

\subsection{Scope and limitations: where current theory stops}
The results above are informative but not exhaustive. Many guarantees rely on convexity, smoothness, or effectively IID sampling; where nonconvex analyses exist, they often target simplified models. Algorithmic stability connects optimization dynamics to generalization and helps explain when faster training can also generalize better \citep{hardt2016stability}, but it does not yet yield tight, predictive statements for modern overparameterized networks across the range of practices used in the field \citep{sun2019optimization,zhang2017understanding}. For this reason, careful \emph{measurement} remains necessary: we should expect theory to suggest which parameters matter (state, order, path), but not to tell us a priori how large their effects will be in a given regime. 
In overparameterized regimes, these stability bounds are often too loose to discriminate neighboring training policies that diverge in function space, so we use stability as a qualitative lens rather than a standalone diagnostic \citep{hardt2016stability,sun2019optimization,zhang2017understanding}.

\noindent\textbf{Implication for this survey.} The families above justify treating optimizer state, sampler state, and path as \emph{separate} sources of training memory, each with its own timescale. They also motivate protocols that (i) perturb one source at a time and (ii) read out effects in function space, not only in loss/accuracy, because noncommutativity and averaging influence what the model \emph{does}, not just what loss it attains.

\section{Empirical Evidence \& Practitioner Heuristics}
\label{sec:empirical}

This section collects recurring empirical findings that make ``training memory'' actionable. We group results by where memory arises (optimizer state, sampler/order, distributed settings, and representation drift), emphasize \emph{what was reported} in the original studies, and extract the simple heuristics practitioners rely on. 

\subsection{Optimizer state effects}
Warm restarts and momentum resets visibly change outcomes even when the model and data are fixed. Cosine schedules with warm restarts improve anytime performance and reach strong accuracy with fewer epochs, illustrating that carrying (or clearing) velocity state across cycles matters \citep{loshchilov2018adamw}. Exponential weight averaging---from classical Polyak averaging to SWA---consistently improves test accuracy and tends to sharpen \emph{calibration} (lower ECE/NLL) by smoothing jagged late-stage trajectories \citep{izmailov2018swa,maddox2019swag}. Sharpness-aware steps (SAM) shift solutions toward flatter neighborhoods and raise test accuracy across CIFAR/ImageNet and fine-tuning regimes, again showing that the per-step memory encoded by the optimizer alters generalization \citep{foret2020sharpness}. Independently, calibration studies report that modern deep nets are often overconfident, and that ensembling/averaging is a strong practical fix \citep{guo2017calibration}.

\textbf{Heuristics used.} (i) When restarting or phase-shifting, explicitly specify momentum/EMA carry-over vs. reset; (ii) SWA/EMA late in training often improves stability and calibration; (iii) Sharpness-aware updates mitigate sharp-minima overfitting and brittle validation curves.

\subsection{Order and curriculum effects}
Order within and across epochs is not an implementation detail. Comparisons of with-replacement vs.\ random reshuffling (RR) show different noise structures; RR often converges faster or to better plateaus even in simple baselines, and practice reflects this preference in large-scale codebases \citep{mishchenko2020random,gurbuzbalaban2021rr}. Pacing policies and curricula stabilize early training and can reduce variance across seeds; surveys document gains across vision and NLP tasks, with improvements measured in accuracy and sample efficiency rather than exotic surrogates \citep{soviany2022curriculum}. In continual settings, small \emph{replay buffers} mitigate catastrophic forgetting and improve average accuracy/backward-transfer metrics, underscoring that sampler memory (what is rehearsed, how often) is decisive \citep{Rebuffi2017iCaRL,LopezPaz2017GEM}.

\textbf{Heuristics used.} (i) Single-epoch RR is generally preferred for SGD; (ii) staged augmentations or curricula can alleviate early instability; (iii) in sequential-task settings, a small, well-chosen replay set reduces forgetting.

\subsection{Distributed/federated effects}
\begin{figure}[t]
  \centering
  \includegraphics[width=0.75\linewidth]{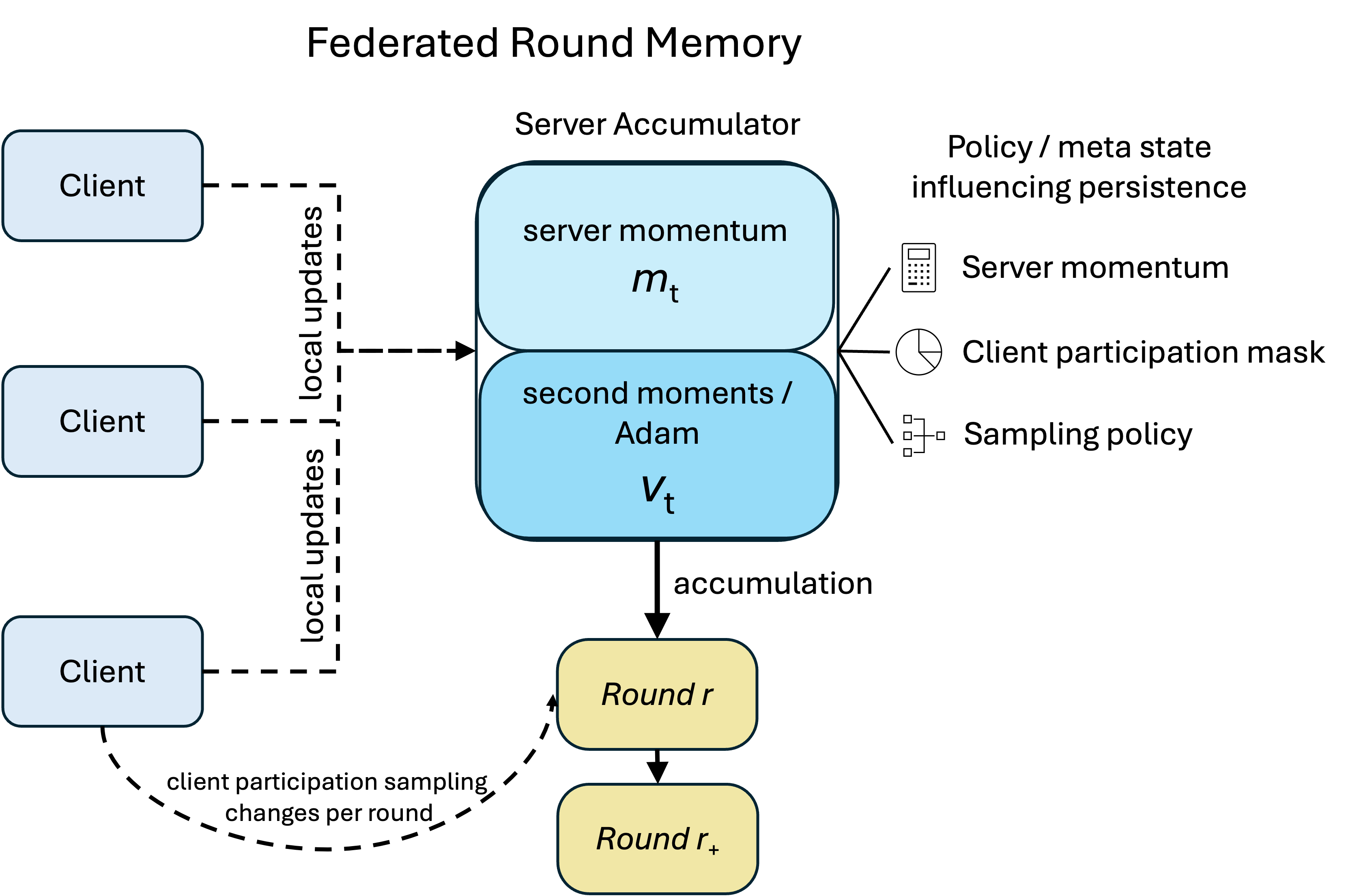}
  \caption{Federated learning rounds with server-side memory: client local updates aggregate into server accumulators (e.g., momentum $m_t$, second moments $v_t$), which persist across rounds; participation masks and sampling policy contribute additional round-scale memory.}
  \label{fig:federated-memory}
\end{figure}

Across rounds, federated optimization exhibits \emph{server-side memory}: FedAvg maintains a running aggregate; adding server momentum/Adam (FedOpt variants) changes both speed and final accuracy, while control-variate methods (e.g., SCAFFOLD) specifically target cross-round \emph{drift} due to heterogeneous clients \citep{mcmahan2017fedavg,reddi2021fedopt,Karimireddy2020Scaffold,Li2020FedProx}. Empirically, reported metrics are validation accuracy (global and per-client), stability across rounds, and fairness measures; sensitivity to client sampling/ordering is a practical concern when participation is sparse or non-IID. A round-level view with server accumulators and client sampling is shown in \figureautorefname~\ref{fig:federated-memory}.

\textbf{Heuristics used.} (i) Server momentum/adaptivity should be logged and tuned; (ii) drift-correction stabilizes training under high heterogeneity; (iii) client sampling/ordering should be documented because it interacts with cross-round memory.

\subsection{Representation drift (what changes, even when accuracy looks fine)}
SVCCA/CKA and related tools show that intermediate representations evolve substantially across epochs, phase boundaries, and runs with different orders or seeds \citep{raghu2017svcca,kornblith2019cka}. Studies typically report \emph{similarity curves} (SVCCA/CKA values over time) alongside accuracy; many also note that lowered similarity across phases does not necessarily imply worse performance, cautioning against causal claims from similarity alone. Still, these measures are informative descriptors of how much history the model ``keeps'' at the feature level.

\textbf{Heuristics used.} A lightweight similarity signal (SVCCA/CKA or stitching) across phases provides context for accuracy changes; interpretation should remain qualitative rather than a stand-in for generalization.

\begin{table}[t]
\centering
\scriptsize
\setlength{\tabcolsep}{3pt}
\renewcommand{\arraystretch}{1.05}
\begin{tabularx}{\linewidth}{l l XX}
\toprule
\textbf{Technique} & \textbf{Setup (typical)} & \textbf{Reported} & \textbf{Empirical takeaway} \\
\midrule
Warm restarts / momentum resets\textsuperscript{\citep{loshchilov2015online}} &
CIFAR/ImageNet; cosine LR with restarts &
Top-1; epochs-to-target &
LR restarts (and velocity handling) alter trajectory; better anytime performance. \\

EMA/SWA weight averaging\textsuperscript{\citep{izmailov2018swa,maddox2019swag}} &
CIFAR/ImageNet; late-phase averaging &
Acc.; ECE; NLL &
Averaging improves generalization; often improves calibration; smoother endpoints. \\

SAM (sharpness-aware)\textsuperscript{\citep{foret2020sharpness}} &
Standard vision benchmarks &
Acc.; robustness &
Flatter-neighborhood solutions generalize better across models/data sets. \\

Order: RR vs.\ replacement\textsuperscript{\citep{mishchenko2020random,gurbuzbalaban2021rr}} &
Supervised image tasks &
Train/val loss; acc. &
RR shows different (often faster) convergence and plateaus than replacement. \\

Curricula/pacing\textsuperscript{\citep{soviany2022curriculum}} &
Vision/NLP; staged difficulty/augs &
Acc.; sample efficiency &
Early stability and sample efficiency improve with pacing. \\

Replay (continual)\textsuperscript{\citep{Rebuffi2017iCaRL,LopezPaz2017GEM}} &
Class-incremental CIFAR/ImageNet &
Avg.\ acc.; BWT/forgetting &
Small rehearsals reduce forgetting; stabilize representation drift. \\

Federated server adaptivity\textsuperscript{\citep{mcmahan2017fedavg,reddi2021fedopt,Karimireddy2020Scaffold,Li2020FedProx}} &
FedAvg/FedOpt variants &
Global acc.; per-round stability &
Server momentum/Adam and drift correction matter under heterogeneity. \\

Representation similarity\textsuperscript{\citep{raghu2017svcca,kornblith2019cka}} &
Across seeds/phases &
SVCCA/CKA curves &
Large representational changes can coexist with similar accuracy; descriptive, not causal. \\
\bottomrule
\end{tabularx}
\caption{Common empirical findings about training memory (\emph{compact}). Metrics are those reported in the cited papers; the rightmost column records the qualitative takeaway as stated by authors.}
\label{tab:empirical-evidence}
\end{table}

\tableautorefname~\ref{tab:empirical-evidence} summarizes recurring empirical patterns, what was reported, and the practical takeaway. Across optimizers, samplers, and distributed settings, \emph{state that persists across steps/epochs/rounds} changes both the path and the endpoint. Practically, the strongest and most portable levers are: (i) make optimizer state handling explicit (carry vs.\ reset), (ii) treat data order and replay as algorithmic choices, and (iii) document cross-round state in federated runs. Representation-similarity tools add context but should be read as descriptors rather than causal attributions.

\section{Measurement Limitations in the Literature}
Training \emph{memory}—history dependence from optimizer state, sampler state, and the path a model takes through parameter space—is widely acknowledged. But when we read closely how papers \emph{measure} its impact, a pattern emerges: several choices that plausibly carry history are changed together, the evaluation lens is narrow, and the procedural details that would enable clean attribution are under-described. The result is that improvements are easy to claim and hard to ascribe.

A first source of confusion is attribution. Modern recipes typically combine momentum or adaptive moments in the optimizer with per-epoch random reshuffling or prioritized sampling in the data pipeline, all while operating in a nonconvex landscape where the path itself can matter. Theory and empirics already tell us that these levers can \emph{independently} alter dynamics—without-replacement (reshuffled) SGD does not behave like with-replacement SGD \citep{shamir2016without,gurbuzbalaban2021rr,mishchenko2020random}. Yet gains are often reported as a single top-line improvement, with little clarity about which parameter did the work. Reproducibility reports have flagged the same issue from another angle: missing information about data order, augmentation stages, or buffer handling makes it difficult to reconstruct what was actually varied \citep{pineau2021reproducibilityreport}.

Ablation studies help, but they are rarely designed as \emph{interventions}. To identify a source of memory, one must vary that source while holding the others fixed. In practice, we seldom see protocols that, for example, reset momentum buffers but keep the exact sampler and order intact, or swap sampler policies while preserving optimizer state. Adjacent fields have shown why this matters: variance across seeds and configurations can flip conclusions and demands statistical controls, not just point estimates \citep{henderson2018matters}. Supervised training papers have adopted some of the reporting hygiene, but true causal contrasts remain uncommon.

\begin{algorithm}[t]
\caption{Branch-and-Hold: single-source interventional diagnostic}
\label{alg:branch-and-hold}
\KwIn{model $\mathcal{M}$ with params $\theta$, optimizer $\mathcal{O}$, sampler policy $\mathcal{S}$, schedule $\Lambda$, 
      probe $\mathcal{P}$, distance $D$ or scalar metric $M$, window $W$, branch time $t_0$, horizon $T$, 
      intervention $\phi$ (perturbs exactly one source), seeds $\mathcal{R}$, log sink $\mathcal{L}$.}
\KwOut{Early effect $\Delta^{\text{early}}$ (at $t_0{+}W$) and final effect $\Delta^{\text{final}}$ (at $T$), each with $95\%$ CI, plus audit logs.}
\BlankLine
\ForEach{seed $r \in \mathcal{R}$}{
  Set global RNG to $r$; record and name RNG streams for: init, sampler/order, augmentation, model\;
  Train a \textbf{root} run to step $t_0$ under $(\mathcal{O},\mathcal{S},\Lambda)$\;
  Persist: $(\theta_{t_0},\text{optimizer buffers}_{t_0},\text{teacher/EMA/SWA}_{t_0},\text{BN stats}_{t_0},\text{sampler state}_{t_0})$\;
  Pre-fetch and store the next $W$ minibatch \emph{IDs} and augmentation RNG states; compute \emph{order hash} for $[t_0,t_0{+}W)$\;
  Deep-copy the full state into two branches: \textsc{Control}, \textsc{Treat}\;

  \tcp{Apply the \emph{single} intended perturbation in Treat at $t_0$; keep all else equal for $W$}
  \textbf{Treat:} apply $\phi$ at $t_0$\;
  \textbf{Both branches:} replay the recorded $W$ minibatch IDs and augmentation RNGs in lockstep; advance schedules identically\;

  \tcp{Early readout at the end of the hold window}
  Evaluate both on probe $\mathcal{P}$; 
  \lIf{$D$ is provided}{set $z_r^{\text{early}} \leftarrow \frac{1}{|\mathcal{P}|}\sum_{x\in\mathcal{P}} D(P_{\text{ctrl}}(x),P_{\text{treat}}(x))$}
  \lElse{set $z_r^{\text{early}} \leftarrow M(f_{\text{treat}})-M(f_{\text{ctrl}})$ on $\mathcal{P}$}

  \tcp{Resume normal training to $T$ with the same high-level policies; RNGs free to evolve}
  Resume standard dataloading from $t_0{+}W$ with identical \emph{policies} $(\mathcal{S},\Lambda)$ in both branches; do not enforce identical micro-order beyond the recorded window\;
  Optionally recalibrate BN before final eval if SWA/EMA is used\;
  Train both branches to horizon $T$; evaluate on $\mathcal{P}$ to get $z_r^{\text{final}}$ via $D$ or $M$ as above\;

  Log to $\mathcal{L}$: order hash, buffer norms (e.g., $\|m\|,\|v\|$), EMA/teacher decay, BN checksum, queue fingerprints (if any)\;
}
Compute $(\Delta^{\text{early}},\mathrm{CI}_{95})$ by calling Alg.~\ref{alg:ate-ci} on $\{z_r^{\text{early}}\}$; same for $\Delta^{\text{final}}$ using $\{z_r^{\text{final}}\}$\;
\KwRet{$\Delta^{\text{early}}$, $\Delta^{\text{final}}$, CIs, and audit logs}
\end{algorithm}

Concretely, a minimal \emph{branch-and-hold} design (Algorithm~\ref{alg:branch-and-hold}) can be instantiated as follows. (i) At a pre-specified step $t$, the full training state $S_t$ is snapshotted—current weights $\theta_t$; optimizer buffers (e.g., momentum/Adam moments); the sampler’s order state (or a record of the index sequence for the next window); and any external or teacher state. (ii) Two branches, $A$ and $B$, are then initialized from this snapshot and differ only in the targeted source (e.g., \textsc{carry} vs.\ \textsc{reset} for optimizer state; \textsc{RR} vs.\ \textsc{WR} for order), with all other ingredients of $\pi$ held constant. (iii) Over the subsequent $W$ updates, data order and randomization streams are matched across branches so that only the intended source is perturbed; training then proceeds to the horizon $T$. Function-space metrics $M(\cdot)$ are evaluated on a fixed probe, and paired ATEs are summarized as in §\ref{subsec:estimands}, with $W$ chosen to reflect the source’s characteristic timescale (e.g., half-life for momentum/EMA, one reshuffled epoch for RR vs.\ replacement, or queue turnover for memory banks).

The measurement lens is also narrow. Accuracy and loss dominate reporting, even though they can obscure calibration errors and instability under shift. A model can improve top-1 while becoming more overconfident \citep{guo2017calibration}, and methods with similar accuracy can diverge sharply in predictive uncertainty when distributional conditions change \citep{ovadia2019can}. Benchmarks that probe subpopulations and domains (e.g., WILDS) repeatedly show uneven behavior that a single aggregate metric hides \citep{koh2021wilds}. Representation-similarity tools such as SVCCA or CKA add welcome visibility into \emph{change} across runs or phases, but the literature cautions that different indices need significance testing or resampling before they support claims about causality \citep{raghu2017svcca,kornblith2019cka,ding2021grounding,davari2023ckareliability}.

Procedure matters as much as metrics, and here too details are thin. Seemingly small choices—whether sampling is with- or without-replacement, how and when augmentations are staged, whether optimizer buffers are carried across warm restarts—alter trajectories in ways both theory and experiments have documented \citep{shamir2016without,gurbuzbalaban2021rr,mishchenko2020random}. Program-level reviews keep finding that such choices are under-specified \citep{pineau2021reproducibilityreport}. Beyond reproducibility concerns, this connects to \emph{underspecification}: pipelines with indistinguishable held-out accuracy can encode very different behaviors under shift \citep{damour2022underspecification}.

Finally, the community underuses early signals and underreports uncertainty. There is rich work on learning-curve and scaling-law extrapolation \citep{swersky2014freezethaw,domhan2015lce,klein2017lcp,kaplan2020scaling,hoffmann2022chinchilla}, but those threads rarely ask whether \emph{early} differences induced by optimizer or sampler state \emph{predict} \emph{late} generalization across policies. When such signals are reported, they often appear without uncertainty, and single-seed results remain common despite evidence that they can reverse conclusions \citep{henderson2018matters,dror2018hitchhikers}. Heterogeneous probe sets and mixed statistical practices further complicate synthesis across papers. A practical default is to run at least five seeds on small/medium benchmarks and at least three on costlier regimes, pairing seeds across branches when estimating $\widehat{\mathrm{ATE}}$. Report the mean and $95\%$ confidence intervals \emph{and} print CI width next to each point estimate; show seed-level scatter when space allows. When asserting “no material difference,” pre-declare an equivalence margin $\varepsilon$ and use an equivalence test rather than relying on overlapping CIs. As a pragmatic convention, resources permitting, one may use at least five seeds on small/medium benchmarks and at least three on costlier regimes, pairing seeds across branches when estimating $\widehat{\mathrm{ATE}}$. Reporting the mean with 95\% confidence intervals—along with the CI width next to each point estimate—and, where space allows, seed-level scatter plots can aid interpretation. Claims of “no material difference” are more defensible when accompanied by a pre-specified equivalence margin $\varepsilon$ and an equivalence procedure (e.g., TOST), rather than relying solely on overlapping confidence intervals.

When we assert “no material difference,” we propose to conduct an \emph{equivalence} test rather than rely on a non-significant difference test. Let $\delta_i$ be the paired, seed-matched effect (e.g., for seed $i$, the mean $\Delta$ under condition B minus condition A) and a practically negligible margin $\varepsilon>0$ in the same units as $\delta_i$ (e.g., total-variation units on the probe). Equivalence is then assessed with the \emph{Two One-Sided Tests} (TOST) procedure on the mean paired effect $\bar\delta$: then reject the composite null $|\bar\delta|\ge\varepsilon$ if and only if both one-sided $t$ tests (lower and upper) are significant at level $\alpha$ with $df=n-1$ degrees of freedom; equivalently, the $(1-2\alpha)$ two-sided confidence interval for $\bar\delta$ lies entirely within $(-\varepsilon,\varepsilon)$. Note that $\delta_i$ needs formed at the seed level (averaging repeats within seed) to avoid pseudoreplication. Alongside the TOST decision report $(\bar\delta, s, n)$, the chosen $\varepsilon$ with rationale, both one-sided $p$-values, and the $(1-2\alpha)$ CI. For small $n$ or uncertain normality, a bootstrap CI check for containment within $(-\varepsilon,\varepsilon)$ can be added. See \citet{schuirmann1987tost,wellek2010equivalence,lakens2017equivalence}.

In short, the literature establishes that history enters through optimizer state, sampler state, and path, and that protocol choices can be causal. What is missing is a \emph{portable, causal, and statistically principled} way to quantify \emph{how much} each source contributes under transparent, reproducible protocols, and to report those effect sizes with uncertainty.

\begin{table}[t]
\centering
\footnotesize
\setlength{\tabcolsep}{3pt}
\renewcommand{\arraystretch}{1.08}
\begin{tabularx}{\linewidth}{l XXX}
\toprule
\textbf{Limitation} & \textbf{Typical manifestation} & \textbf{Why it hinders attribution} & \textbf{What to control/report} \\
\midrule
Attribution ambiguity &
Change optimizer \emph{and} sampler; report one headline gain &
Multiple stateful sources move; effects mixed &
Hold two sources fixed, perturb one; log carry-over of buffers across phases \\
\addlinespace[2pt]
Causal gap &
Ablations without explicit resets or swaps &
No interventional contrast to isolate optimizer vs.\ sampler vs.\ path &
Include reset/swap interventions; pre-declare attribution tests and seeds \\
\addlinespace[2pt]
Metric myopia &
Only top-1/loss curves are reported &
Calibration, OOD reliability, and function-space movement are hidden; rep-metrics lack uncertainty &
Add calibration/shift stress tests; function-/rep-space deltas with CIs; equivalence tests when claiming ``no difference'' \\
\addlinespace[2pt]
Under-specification &
Sampler policy, augmentation schedule, buffer handling omitted &
Pipelines look ``equivalent'' in accuracy but behave differently under shift &
Specify order policy, augmentation phases, and state handling at restarts; share configs and seeds \\
\addlinespace[2pt]
Early-phase blindness &
No early leading indicators &
Predictive signal about later generalization is unused &
Report early-window indicators (with uncertainty) and correlate with final outcomes across policies \\
\addlinespace[2pt]
Reporting heterogeneity &
Single seed; no CIs or statistical tests &
High variance can flip conclusions; hard to synthesize across papers &
Multi-seed summaries; paired/bootstrap CIs; appropriate tests and declared equivalence margins \\
\bottomrule
\end{tabularx}
\caption{Failure modes that obscure measurement of training memory and pragmatic controls to make results interpretable—without prescribing a specific diagnostic.}
\label{tab:gap-failures}
\end{table}

\tableautorefname~\ref{tab:gap-failures} summarizes common failure modes and pragmatic controls. These failure modes share a common root: there is lack of \emph{explicit targets of estimation} that isolate optimizer state, sampler policy,
or path effects while holding everything else fixed. To make attribution
portable and testable, we now formalize source-specific \emph{causal estimands}
that every diagnostic should report. These estimands turn the informal notion of
“training memory matters” into concrete, seed-averaged effect sizes with
uncertainty.

\subsection{Causal Estimands for Attribution}
\label{subsec:estimands}
\begin{figure*}[t]
  \centering
  \includegraphics[width=\textwidth]{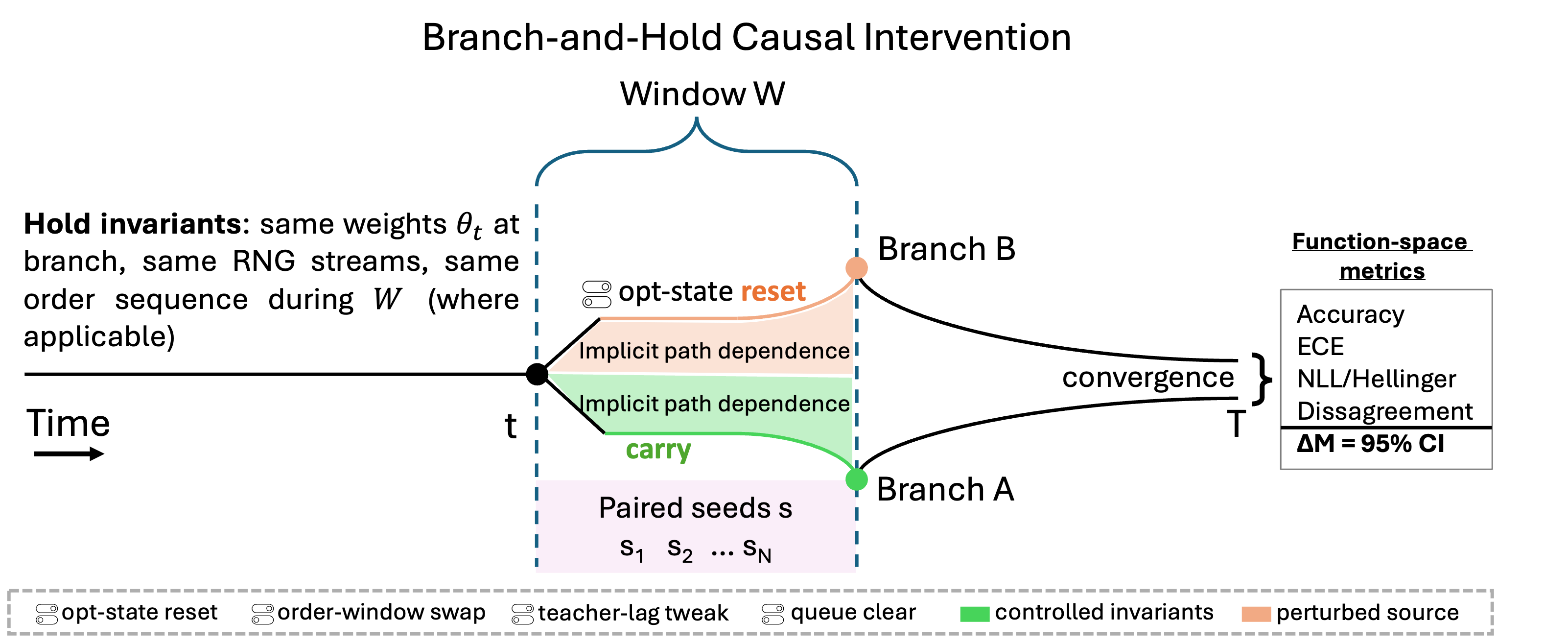}
  \caption{Branch-and-hold design at time $t$: fork runs that differ only in a targeted source (e.g., \textsc{carry} vs.\ \textsc{reset} optimizer state, or order window swap) for a window $W$, then continue to horizon $T$. Report paired seed effect sizes $\Delta M$ in function space with CIs.}
  \label{fig:branch-hold}
\end{figure*}

Fix a training recipe $\pi$ (architecture, optimizer hyperparameters,
augmentation, evaluation protocol) and a horizon $T$. Let $f_T^{(i,s)}$ denote
the predictor obtained after $T$ updates when we apply intervention $i$ and
random seed $s$, with \emph{all other} ingredients of $\pi$ held fixed. Let
$M(\cdot)$ be a scalar \emph{function-space} metric on a fixed evaluation
distribution $\mathcal{D}_{\mathrm{eval}}$ (e.g., accuracy $\uparrow$, ECE/NLL
$\downarrow$, or pairwise disagreement on a probe set). We define source-specific
average treatment effects (ATEs) as expectations over seeds:
\[
\mathrm{ATE}_{\mathrm{opt}}
= \mathbb{E}_{s}\!\left[\, M\!\left(f_T^{(\textsc{carry},\,s)}\right) -
M\!\left(f_T^{(\textsc{reset},\,s)}\right) \right],\qquad
\mathrm{ATE}_{\mathrm{order}}
= \mathbb{E}_{s}\!\left[\, M\!\left(f_T^{(\textsc{RR},\,s)}\right) -
M\!\left(f_T^{(\textsc{WR},\,s)}\right) \right].
\]
Here, \textsc{carry}/\textsc{reset} differ only in optimizer-state handling at a
predeclared branch point (e.g., momentum/Adam buffers, EMA), while
\textsc{RR}/\textsc{WR} differ only in sampling policy (random reshuffling vs.\
with-replacement). Finite-sample estimates use $S$ \emph{paired} seeds,
$\widehat{\mathrm{ATE}}=\tfrac{1}{S}\sum_{j=1}^{S}\!\big[M(f_T^{(i_1,s_j)})-
M(f_T^{(i_0,s_j)})\big]$, with $95\%$ bootstrap confidence intervals. The same
pattern yields $\mathrm{ATE}_{\mathrm{teacher}}$ (EMA decay $\alpha\!\to\!\alpha'$),
$\mathrm{ATE}_{\mathrm{queue}}$ (contrastive-memory clear vs.\ carry), or
\emph{order-window} ATEs where only a window of $W$ minibatches is permuted. Our estimator and uncertainty summary are given in Algorithm~\ref{alg:ate-ci}, and the branching design is illustrated in \figureautorefname~\ref{fig:branch-hold}.

\begin{algorithm}[t]
\caption{Paired ATE and Bootstrap CI}
\label{alg:ate-ci}
\KwIn{Per-seed paired measurements $\{z_r\}_{r\in\mathcal{R}}$ (e.g., mean probe distance or metric delta for seed $r$); bootstrap rounds $B$.}
\KwOut{Point estimate $\widehat{\mathrm{ATE}}$ and $95\%$ CI.}
\BlankLine
Compute $\widehat{\mathrm{ATE}} \leftarrow \frac{1}{|\mathcal{R}|}\sum_{r\in\mathcal{R}} z_r$\;
\For{$b \gets 1$ \KwTo $B$}{
  Sample with replacement a multiset $\mathcal{R}^{(b)}$ from $\mathcal{R}$\;
  Set $\widehat{\mathrm{ATE}}^{(b)} \leftarrow \frac{1}{|\mathcal{R}^{(b)}|}\sum_{r\in\mathcal{R}^{(b)}} z_r$\;
}
Let $\mathrm{CI}_{95}$ be the 2.5th–97.5th percentiles of $\{\widehat{\mathrm{ATE}}^{(b)}\}_{b=1}^B$\;
\KwRet{$\widehat{\mathrm{ATE}}$, $\mathrm{CI}_{95}$}
\end{algorithm}

\subsection{Portable Interventions (Perturbation Primitives)}
\label{subsec:primitives}
The estimands in §\ref{subsec:estimands} require targeted perturbations that
change exactly one source of training memory while holding others fixed.
\tableautorefname~\ref{tab:perturbation-primitives} lists minimal, portable interventions
that map directly to the taxonomy in §\ref{sec:taxonomy} (S1–S5) and to typical
lifetimes (step, epoch, phase, task). They are architecture- and dataset-agnostic
and pair naturally with paired-seed ATEs and bootstrap confidence intervals. We implement these single-source perturbations via Algorithm~\ref{alg:opt-reset} (optimizer-state reset), Algorithm~\ref{alg:order-swap} (order-window swap), Algorithm~\ref{alg:phase-policy} (phase-boundary policy), and Algorithm~\ref{alg:queue} (external-memory interventions).

\begin{algorithm}[t]
\caption{Optimizer-state reset (S1): zero/rewarm buffers; sampler fixed for $W$}
\label{alg:opt-reset}
\KwIn{As in Alg.~\ref{alg:branch-and-hold}; momentum/AdamW hyperparams $(\beta_1,\beta_2)$; optional EMA/SWA handle; optional LR rewarm length $K$.}
\KwOut{$\Delta_{\text{opt}}^{\text{early}}$, $\Delta_{\text{opt}}^{\text{final}}$ + CIs; buffer-state provenance.}
\BlankLine
Define $\phi_{\text{opt}}$ at $t_0$:
\begin{itemize}
  \item For SGD+momentum: set velocity to zero.
  \item For Adam/AdamW: zero first and second moments; keep weight parameters identical; preserve weight decay state (decoupled).
  \item Optionally pause EMA/SWA updates for the next $W$ steps (freeze teacher / averaging buffers).
  \item If using LR rewarm, apply the same $K$-step warmup schedule in \emph{both} branches to avoid confounding.
\end{itemize}
Choose $W \approx 1$--$2$ half-lives where $h_{1/2}=\log(0.5)/\log(\beta_1)$ for momentum/Adam first moment; note Adam’s second moment has half-life $\log(0.5)/\log(\beta_2)$\;
Call Alg.~\ref{alg:branch-and-hold} with $\phi \leftarrow \phi_{\text{opt}}$, enforcing identical minibatch IDs and augmentation RNGs only on $[t_0,t_0{+}W)$\;
\KwRet{$\Delta_{\text{opt}}^{\text{early}}$, $\Delta_{\text{opt}}^{\text{final}}$ + CIs}
\end{algorithm}

\begin{algorithm}[t]
\caption{Order-window swap (S2): permute a recorded window; optimizer state fixed}
\label{alg:order-swap}
\KwIn{As in Alg.~\ref{alg:branch-and-hold}; window size $W$ (one epoch for RR; fixed steps for WR).}
\KwOut{$\Delta_{\text{order}}^{\text{early}}$, $\Delta_{\text{order}}^{\text{final}}$ + CIs; order/augmentation hashes.}
\BlankLine
At $t_0$, record the next $W$ minibatch \emph{IDs} and augmentation RNG states under policy $\mathcal{S}$; compute order and augmentation hashes\;
Define $\phi_{\text{order}}$:
\begin{itemize}
  \item \textsc{Control}: replay recorded order and augmentation RNGs for the window.
  \item \textsc{Treat}: replay the same \emph{multiset} of $W$ minibatches under a fresh permutation; reuse the recorded augmentation RNG per-example (or re-seed deterministically to isolate \emph{order} only).
\end{itemize}
Ensure optimizer buffers, EMA/SWA, BN stats, and schedule $\Lambda$ are identical at $t_0$ across branches\;
Call Alg.~\ref{alg:branch-and-hold} with $\phi \leftarrow \phi_{\text{order}}$\;
\KwRet{$\Delta_{\text{order}}^{\text{early}}$, $\Delta_{\text{order}}^{\text{final}}$ + CIs}
\end{algorithm}

The intervention window $W$ is the horizon over which a perturbation is applied (or held)
so that its influence is visible but not obscured by unrelated drift. A good
rule is to match $W$ to the characteristic \emph{memory} of the source that is being probed. For momentum/Adam, the relevant timescale is the half-life of the
exponential smoothing, $h_{1/2}=\log(0.5)/\log(\beta)$: with $\beta{=}0.9$ the
signal halves in roughly $6\!-\!7$ steps, while $\beta{=}0.99$ stretches this
to about $69$ steps (and $\beta{=}0.999$ to hundreds). Choosing
$W\!\approx\!1$–$2$ half-lives typically balances power and interference: too
short and the effect is underresolved; too long and other factors creep in.
For order interventions, portability argues for one full epoch under random
reshuffling, or a fixed minibatch window when sampling with replacement.
Teacher–student EMA follows an analogous logic: the effective averaging horizon
is on the order of $1/(1-\alpha)$, so $\alpha{=}0.99$ exposes change over
$\sim 10^2$ steps and $\alpha{=}0.999$ over $\sim 10^3$. External memory
(e.g., contrastive queues) suggests $W$ comparable to the queue’s turnover
(queue size divided by enqueue rate), long enough for the bank to “forget”
past contents without permanently altering the training distribution.

\begin{algorithm}[t]
\caption{Phase-boundary policy (S1): carry/reset/rewarm at pretrain$\rightarrow$finetune}
\label{alg:phase-policy}
\KwIn{Checkpoint $(\theta^\star,\mathcal{O}^\star,\text{EMA/SWA}^\star,\text{BN}^\star)$ from pretraining; finetune data; options \{\textsc{carry}, \textsc{reset}, \textsc{rewarm}\}; probe $\mathcal{P}$; early window size $k$ epochs.}
\KwOut{Early and final effects $\Delta_{\text{phase}}$ with CIs; calibration deltas.}
\BlankLine
Construct three branches at the phase start $t_0$:
\begin{enumerate}
  \item \textsc{Carry}: load $(\theta^\star,\mathcal{O}^\star,\text{EMA/SWA}^\star,\text{BN}^\star)$ unchanged.
  \item \textsc{Reset}: load $\theta^\star$; zero optimizer moments (SGD velocity; Adam $m,v$); reinit EMA/SWA and BN running stats.
  \item \textsc{Rewarm}: as \textsc{Reset}, plus $K$-step LR warmup (same $\Lambda$ thereafter).
\end{enumerate}
Hold sampler policy, seeds, and augmentation pipelines identical across branches; match target LR at the phase start\;
Compute early effects over the first $k$ epochs; continue to convergence; report accuracy and calibration (ECE/NLL) alongside function-space $\Delta$\;
\KwRet{$\{\Delta_{\text{phase}}^{\text{early}}, \Delta_{\text{phase}}^{\text{final}}\}$ + CIs}
\end{algorithm}

\begin{algorithm}[t]
\caption{External-memory intervention (S4): freeze or clear queue/bank for $W$}
\label{alg:queue}
\KwIn{Queue/bank size $K$; enqueue rate $m$ per step; window $W \approx K/m$; probe $\mathcal{P}$.}
\KwOut{$\Delta_{\text{queue}}^{\text{early}}$, $\Delta_{\text{queue}}^{\text{final}}$ + CIs; queue fingerprints.}
\BlankLine
Define treatments at $t_0$:
\begin{enumerate}
  \item \textsc{Freeze}: stop enqueue/dequeue for $W$ steps (read-only snapshot).
  \item \textsc{Clear}: empty queue, then repopulate under normal policy (stronger perturbation).
\end{enumerate}
Hold optimizer state, sampler order for the window, and schedule identical across branches; hash queue contents at $t_0$ and $t_0{+}W$\;
Invoke Alg.~\ref{alg:branch-and-hold} with $\phi \in \{\textsc{Freeze}, \textsc{Clear}\}$\;
\KwRet{$\Delta_{\text{queue}}^{\text{early}}$, $\Delta_{\text{queue}}^{\text{final}}$ + CIs; note stability differences between treatments.}
\end{algorithm}

\begin{table}[t]
\centering
\scriptsize
\setlength{\tabcolsep}{3pt}
\renewcommand{\arraystretch}{1.07}
\begin{tabularx}{\linewidth}{l X l l}
\toprule
\textbf{Target (S\#)} & \textbf{Intervention (what you change; \emph{hold fixed})} & \textbf{Window $W$} & \textbf{Estimand} \\
\midrule
S1 Optimizer / trajectory &
\textbf{Opt-state reset}: at a predeclared step/epoch, zero momentum/Adam moments; optionally freeze EMA/SWA for $W$ steps \emph{(hold: weights, LR schedule, sampler policy \& seeds, augmentations)} &
$\approx$1–2 half-lives ($h_{1/2}=\log 0.5/\log\beta$) &
$\mathrm{ATE}_{\text{opt}}$ \\
\addlinespace[2pt]
S2 Sampler / data order &
\textbf{Order-window swap}: record the next $W$ minibatches (IDs), replay them once under a fresh permutation, then resume \emph{(hold: weights, optimizer state, LR schedule, augmentation RNG streams)} &
1 epoch (RR) or a fixed window ($10^3$–$2{\times}10^3$ steps) &
$\mathrm{ATE}_{\text{order}}$ \\
\addlinespace[2pt]
S1{+}S4 Phase boundary &
\textbf{Phase policy}: at checkpoint (e.g., pretrain$\!\to\!$finetune), branch into \textsc{carry}/\textsc{reset}/\textsc{rewarm} for momentum/Adam and EMA/SWA \emph{(hold: data order, seeds, eval protocol; equal target LR at phase start)} &
First $k$ epochs of the new phase ($k{=}1$–$5$) &
$\mathrm{ATE}_{\text{opt}}$ (phase) \\
\addlinespace[2pt]
S5 Meta-state (teacher) &
\textbf{Teacher-lag tweak}: temporarily change teacher EMA decay $\alpha\!\to\!\alpha'$ for $W$ steps, then restore \emph{(hold: student optimizer state, sampler policy, weights at branch)} &
$\approx c/(1-\alpha)$ for small $(1-\alpha)$ &
$\mathrm{ATE}_{\text{teacher}}$ \\
\addlinespace[2pt]
S4 External memory &
\textbf{External-memory ablation}: clear or \emph{freeze} contrastive queues/memory banks for $W$ steps (no enqueue/dequeue), then resume \emph{(hold: optimizer state, sampler order, LR schedule, seeds)} &
$\approx$ queue length / enqueue rate &
$\mathrm{ATE}_{\text{queue}}$ \\
\bottomrule
\end{tabularx}
\caption{Minimal interventions for source-specific attribution of training memory. Each perturbs exactly one source, uses paired seeds, and is summarized with seed-averaged effect sizes and bootstrap CIs (cf.\ §\ref{subsec:estimands}).}
\label{tab:perturbation-primitives}
\end{table}

Causal contrasts live or die by the guarantee that the two branches differ
\emph{only} in the intended perturbation. In practice, this can be facilitated by adopting a single root seed and deriving separate, documented streams for the sampler/order, augmentations, and model-side randomness; persisting the exact sequence of example indices (e.g., per-epoch “order hashes”) allows order to be replayed verbatim. Potential sources of nondeterminism may be constrained by fixing framework and toolkit versions, enabling deterministic kernels where available (e.g., cuDNN determinism flags), and pinning dataloader parameters (number of workers, prefetching, sharding), while being mindful of operators that are nondeterministic on the target hardware. At branch points, snapshotting and logging optimizer buffers and any teacher/EMA or external-memory state makes \textsc{carry} and \textsc{reset} policies auditable. Taken together, these controls (see \tableautorefname~\ref{tab:isolation-summary}) help isolate the intervention and align with the protocol outlined in §\ref{sec:benchmarks}.

\begin{table}[t]
\centering
\small
\setlength{\tabcolsep}{4pt}
\renewcommand{\arraystretch}{1.08}
\begin{tabularx}{\linewidth}{l X X}
\toprule
\textbf{Control} & \textbf{Practice} & \textbf{Purpose} \\
\midrule
Randomness contract &
Identify all randomness sources (initialization, data order, augmentation, batching, parallel execution). Derive named streams from a single root key per run and record those keys. &
Reproducibility; enables paired A/B branches with identical noise. \\
\addlinespace[2pt]
Freeze the stack &
Fix code, configuration, and numerical options into an immutable artifact; record dependency manifests and a lightweight hardware fingerprint. &
Prevents environment drift from masquerading as treatment effects. \\
\addlinespace[2pt]
Persist data order &
Store the exact sequence of example identifiers per step/epoch; emit a compact per-epoch “order checksum’’ for quick verification. &
Allows verbatim replay and equality checks across branches. \\
\addlinespace[2pt]
Snapshot hidden state at branch points &
Save weights and all state that can carry history (optimizer summaries, teacher/EMA or weight-averaging state, external queues/banks, scheduler and sampler state). Declare the policy (\textsc{carry}/\textsc{reset}/\textsc{rewarm}) for each component. &
Makes the intervention auditable; ensures only the intended source differs. \\
\addlinespace[2pt]
Hold invariants across branches &
Keep batch boundaries, order policy, augmentation pipeline, schedules, and parallelism layout identical. Validate post-branch invariants (matching order checksums, schedule values, buffer-norm sanity). &
Isolates the perturbation; reduces interference. \\
\addlinespace[2pt]
Account for residual nondeterminism &
When exact determinism is unattainable, use paired seeds and repeated runs; report paired/bootstrap confidence intervals and CI widths. &
Quantifies remaining noise; guards against overclaiming. \\
\bottomrule
\end{tabularx}
\caption{Isolation and determinism controls (summary). Platform-agnostic practices that ensure branches differ only in the intended perturbation; complements the protocol guidance in §\ref{sec:benchmarks}.}
\label{tab:isolation-summary}
\end{table}

\section{Desiderata for Future Diagnostics}
\label{sec:desiderata}

A useful diagnostic should not merely confirm that training has memory; it should separate \emph{where} that memory comes from, quantify \emph{how much} it matters for what the model does, and do so \emph{early} and \emph{reproducibly}. The points below state properties a diagnostic ought to have, without prescribing a particular method.

\paragraph{D1. Attribution.}
A useful diagnostic isolates optimizer state, sampler/data-order state, and path/geometry as distinct sources of history. The organizing principle is orthogonal perturbation: one source is altered while the others are held fixed and the control is verified. Examples include resetting momentum/Adam buffers at a phase boundary while keeping weights and the exact data order unchanged; freezing the sampler policy across runs while varying whether EMA/SWA is carried over; or swapping the order of two short segments (\emph{A then B} vs.\ \emph{B then A}) to expose non-commutativity. Perturbations are minimal in duration and scope and are chosen on the appropriate timescale (step, epoch, phase, task) so that observed effects are attributable to the intended source rather than collateral changes. Portable interventions that operationalize these single-source perturbations appear in §\ref{subsec:primitives} and \tableautorefname~\ref{tab:perturbation-primitives}.

\paragraph{D2. Function-space sensitivity.}
Differences in \emph{function space}—the model’s predictive distributions or outputs on a held-out \emph{probe set}—are reported alongside accuracy and loss. Task-aligned distances (e.g., total variation, Jensen–Shannon, Hellinger for classification; calibrated error measures for regression/probabilistic models) provide interpretable signals even when top-line accuracy is unchanged. Effect sizes are summarized with uncertainty (e.g., bootstrap confidence intervals), and, where appropriate, equivalence tests support ``no material difference’’ claims rather than relying on point estimates.

\paragraph{D3. Representation tracking.}
Representation-level readouts (e.g., SVCCA/CKA or stitching-based probes) are used to contextualize feature drift across phases or interventions and are treated as descriptive rather than causal. Layers/splits are predefined to avoid cherry-picking. Similarity curves are accompanied by stability checks (resampled probes, noise injections) and are interpreted alongside function-space effects so that disagreements are informative rather than misleading.

\paragraph{D4. Early predictivity.}
Leading indicators are measured during early epochs to forecast final generalization under varied momentum schedules, sampler policies, or phase decisions. Emphasis is on cross-policy predictivity—signals that maintain rank or calibration across the interventions under study—rather than tuned heuristics for a single setting. Predictivity is quantified (e.g., rank correlation, calibrated regression) with uncertainty, and failures of early indicators are reported explicitly.

\paragraph{D5. Protocol clarity.}
Portability is achieved by specifying datasets and model families; random seeds and run counts; the \emph{exact} data-order policy (with- or without-replacement, reshuffle rule and seed); augmentation regimes and their phase changes; optimizer-state handling across epochs/phases (carry vs.\ reset for momentum/Adam/EMA/SWA); probe construction and reuse across interventions; compute budget; and the full statistical treatment (confidence intervals, tests, equivalence margins). Parameters are predeclared where feasible to limit degrees of freedom. The objective is that another lab can replay the same memory perturbations and obtain comparable effect sizes.

\begin{table}[t]
\centering
\small
\renewcommand{\arraystretch}{1.15}
\begin{tabularx}{\linewidth}{l X X}
\toprule
\textbf{Desideratum} & \textbf{Minimal properties to satisfy} & \textbf{Anti-patterns to avoid} \\
\midrule
\textbf{D1 Attribution} &
Single-source perturbations; verify controls (buffer norms, order hashes); choose perturbation timescale to match source &
Changing multiple sources at once; unverified “we reset X”; long perturbations that obscure effects \\
\textbf{D2 Function-space} &
Task-aligned distances on a fixed probe; effect sizes with CIs and (when relevant) equivalence tests &
Only reporting accuracy/loss; distances without uncertainty; ad-hoc probe selection \\
\textbf{D3 Representation} &
Predefined layers/splits; pair similarity with function-space results; stability checks via resampling/noise &
Treating similarity as causal; cherry-picked layers or runs; no robustness analysis \\
\textbf{D4 Early predictivity} &
Early-window measures tested across policies; rank/fit reported with CIs; failures documented &
Tuning on one policy and claiming universality; reporting single numbers without uncertainty \\
\textbf{D5 Protocol clarity} &
Explicit sampler policy, augmentation schedule, state handling across phases; seeds/runs; reusable configs &
Omitted order/augmentation details; unclear buffer handling at restarts; single-seed reports \\
\bottomrule
\end{tabularx}
\caption{Solution-agnostic checklist for diagnostics that measure training memory in a way that is attributable, sensitive in function space, representation-aware, predictive early, and reproducible.}
\label{tab:desiderata}
\end{table}

\section{Benchmarks \& Reporting Checklist}
\label{sec:benchmarks}

The persuasiveness of a diagnostic depends on the testbeds that host it. Selection is guided by three criteria. \textbf{Accessibility:} runs complete on commodity hardware with multiple seeds and interventions (\S\ref{sec:desiderata}), enabling uncertainty reporting beyond single seeds. \textbf{Controllability:} each testbed exposes the parameters that create memory—optimizer state across phase boundaries, sampler policies and order, augmentation or preprocessing stages—so that attribution is meaningful. \textbf{Sensitivity:} the task reacts to step-, epoch-, and phase-scale perturbations; if order or state changes never move the needle, the testbed is uninformative for this purpose.

One encoder family per modality typically suffices; horizons are capped to prioritize seeds and probes. A single well-instrumented setting per modality is more informative than a broad but under-specified suite. Small, controllable tasks are preferred so that single-source perturbations, paired seeds, and probe-based readouts are feasible. \emph{Compact exemplars} that satisfy these constraints are summarized in \tableautorefname~\ref{tab:compact-testbeds}, together with the memory sources they expose and convenient intervention windows~$W$.

\begin{table}[t]
\centering
\small
\renewcommand{\arraystretch}{1.05}
\begin{tabularx}{\linewidth}{l l X}
\toprule
\textbf{Modality} & \textbf{Compact testbed} & \textbf{Memory sources exposed \& convenient perturbations ($W$)} \\
\midrule
Vision & CIFAR-10/100 + ResNet-18/MobileNetV2 & Epoch-scale order (RR vs.\ WR; $W{=}$1 epoch), step-scale optimizer (momentum/Adam/EMA; $W{\approx}$1–2 half-lives), phase boundary (pretrain$\!\to$finetune; first $k$ epochs), staged augs. \\
Language & SST-2/AGNews/IMDb + DistilBERT & Length bucketing/packing (order), tokenization choices (phase), AdamW moments/EMA (step); probe on a fixed corpus slice. \\
Audio & SpeechCommands/ESC-50 + small CNN & Preprocessing stages (STFT/mel) as phase changes; clip order and speaker balancing; accuracy saturates early $\Rightarrow$ calibration \& function-space distances. \\
Graphs & Cora/CiteSeer/PubMed & Neighborhood sampling (order), train/val/test edge splits; schedules/regularizers across phases; fixed node/edge probe. \\
Recsys & MovieLens-100K/1M + MF/NeuMF & Negative-sampling policy (sampler memory), refresh cadence; metrics flat while predictive distribution shifts $\Rightarrow$ probe-based distances. \\
Time series & ETTh1/h2, small M4 splits & Sliding-window construction and scaling (phase), EMA/momentum (step); evaluate forecast distributions and interval coverage. \\
RL & CartPole/MountainCar/A2C; DQN & Replay size/priority (sampler memory), target-network lag (server-like memory); episode order/environment seeds as data-order policy. \\
Federated & FEMNIST/Shakespeare; CIFAR/AGNews (Dirichlet) & Client sampling, server momentum/Adam (FedOpt), control variates; per-client dispersion and probe-based distances on fixed clients. \\
\bottomrule
\end{tabularx}
\caption{Compact testbeds that expose step/epoch/phase/round memory with modest compute. $W$ denotes the intervention window suggested by the source’s characteristic lifetime.}
\label{tab:compact-testbeds}
\end{table}

The \emph{minimum information} required to enable attribution and replay is listed in \tableautorefname~\ref{tab:reporting-checklist}. Items are domain-agnostic; modality-specific notes can be appended as needed. Multiple seeds are used where feasible; summaries include a central tendency (e.g., mean) and uncertainty (e.g., 95\% CI). Paired designs are preferred; uncertainty is estimated via paired resampling/bootstrap over the probe and/or seeds, or via analytic intervals when appropriate. For ``no material difference'' claims, a practical equivalence margin $\varepsilon$ is predeclared and an equivalence test (e.g., TOST) is applied; otherwise, confidence intervals are reported without dichotomous accept/reject language. When many interventions are tested, a false discovery rate is controlled or the number of tests is disclosed, avoiding selective reporting.

\begin{table}[t]
\centering
\small
\renewcommand{\arraystretch}{1.05}
\begin{tabularx}{\linewidth}{l X}
\toprule
\textbf{Category} & \textbf{Specification (to be recorded in paper or artifact)} \\
\midrule
Datasets & Name and version/split; resizing/filtering; federated partition scheme (if applicable). \\
Architectures & Model family/size; heads/tokenizer (NLP); feature front-end (audio). \\
Seeds \& randomness & Initialization, data-order, augmentation/probe RNG; environment/client seeds (RL/FL); number of runs; naming of multiple RNG streams. \\
Sampler policy & With- vs.\ without-replacement and reshuffle rule/seed; bucketing/packing; negative sampling; neighborhood sampling; client sampling. \\
Optimizer \& meta-state & Optimizer family; momentum/EMA/SWA/teacher; carry vs.\ reset at each boundary; BN handling (recalibration if applicable). \\
Schedules & LR schedule and parameters; warmup/restarts; interaction with state (e.g., whether restarts reset or carry buffers). \\
Transforms / preprocessing & Vision augs; tokenization/truncation; STFT/mel; normalization/windowing; whether stages change over time. \\
Compute budget & Max epochs/steps/episodes; batch size; gradient accumulation; evaluation cadence; hardware (context only). \\
Probe (if used) & Size; construction; reuse policy; whether fixed across branches; domain specifics (states for RL, clients for FL). \\
Metrics & Task metrics; at least one calibration/probabilistic metric when relevant; function-/representation-space distances; effect-size definitions. \\
Uncertainty & CI method (e.g., bootstrap with $B$ resamples); equivalence margin $\varepsilon$ for ``no material difference''; multiple-comparison policy. \\
Artifacts & Configs (e.g., YAML); order hashes; buffer/BN-state logs; code commit/containers; evaluation scripts. \\
\bottomrule
\end{tabularx}
\caption{Memory-sensitive reporting checklist (minimum information to enable attribution and replay).}
\label{tab:reporting-checklist}
\end{table}

\section{Related Areas \& Transferable Insights}

Training memory does not live only inside the supervised learning loop. Several nearby communities—continual learning, curriculum and pacing, data selection and coresets, reinforcement learning with replay, and federated optimization—have been explicitly managing (or struggling with) state that persists across updates for years. Reading these areas through the lens of our taxonomy (source, lifetime, visibility) yields concrete lessons for how to \emph{measure} memory in standard deep learning pipelines.

\subsection{Continual learning: explicit long-term memory and its costs.}
Continual learning makes memory the main character: the system must acquire new tasks without catastrophically overwriting old ones. The classic picture is that naive fine-tuning destroys previously learned competencies, motivating mechanisms that \emph{stabilize} parts of the model or \emph{rehearse} past data \citep{parisi2019continual,delange2021survey}. Elastic Weight Consolidation (EWC) penalizes movement along parameters deemed important to prior tasks using a Fisher-based quadratic, an explicit, inspectable memory of past curvature that persists over task scales \citep{kirkpatrick2017overcoming}. Gradient Episodic Memory (GEM) stores exemplars and enforces per-step constraints to avoid increasing loss on earlier tasks, turning a replay buffer into a causal lever on forgetting \citep{LopezPaz2017GEM}. iCaRL mixes exemplar replay with prototype-based classification, highlighting that the \emph{form} of memory (raw samples vs.\ prototypes) changes both compute and measurement \citep{Rebuffi2017iCaRL}. Large surveys synthesize these families—regularization, replay, parameter isolation/expansion—and repeatedly warn that reported gains depend strongly on protocol details: task ordering, buffer budgets, and how optimizer state is handled across task boundaries \citep{parisi2019continual,delange2021survey,Mundt2023Wholistic}. For our purposes: (i) rehearsal-style methods provide ready-made interventions for attribution (toggle, size, and sampling within buffers); (ii) quadratic anchoring (EWC/L2-SP) shows how “implicit” path memory can be made explicit via auxiliary state; and (iii) the community’s metrics (backward/forward transfer, forgetting curves) are exemplars of task-aligned evaluation with uncertainty.

\subsection{Curriculum and pacing: order as a first-class control parameter.}
Curriculum learning and self-paced variants turn data order into policy: start with “easy” examples or high-confidence regions, then expand difficulty \citep{bengio2009curriculum,Kumar2010SelfPaced}. Modern surveys catalog manual and automatic curricula across vision, NLP, and RL, including progress signals, teacher–student schemes, and environment-generated pacing \citep{soviany2022curriculum,Narvekar2020JMLR}. The central insight is that reordering the same multiset of examples can change stability, sample efficiency, and final generalization—an \emph{epoch-scale} memory whose lifetime outlasts transient optimizer noise. That community also surface practical pitfalls that map directly to our gap analysis: curricula are often under-specified (how is “difficulty” scored? how often is the schedule recomputed?), and ablations rarely separate sampler effects from optimizer state (e.g., momentum buffers that retain pre-curriculum statistics). For diagnostics, curricula offer testbeds where order is deliberately structured; causal designs can swap curricula while freezing optimizer state (or vice versa) and measure function-space deltas with confidence intervals.

\subsection{Data selection \& coresets: sampler memory as principled selection.}
A rich line of work formalizes sample \emph{prioritization} and subset selection. Influence functions and data valuation quantify how training points move parameters and predictions, revealing that not all points are equally useful and that their value depends on the current state—an inherently stateful notion of sampling \citep{KohLiang2017Influencea,GhorbaniZou2019DataShapley}. Empirically, example “forgetting events” and early-loss proxies (e.g., EL2N) show that example hardness and learnability evolve across training \citep{Toneva2019Forgetting,Paul2021EL2N}. On the \emph{algorithmic} side, importance sampling for deep nets proposes loss- or gradient-magnitude–weighted sampling to reduce variance and accelerate optimization \citep{katharopoulos2018importance,johnson2018training,loshchilov2015online}. Coreset methods such as CRAIG (gradient matching via submodular surrogates), GLISTER (bi-level generalization-driven selection), and GRAD-MATCH (explicit gradient matching) make the sampler’s state visible and controllable through selected subsets \citep{Mirzasoleiman2020CRAIG,Killamsetty2020GLISTER,Killamsetty2021GradMatch}. These works provide \emph{measurement} scaffolding: gradient-matching gaps, validation lift vs.\ subset size, and per-iteration selection diagnostics. For our study, they suggest straightforward interventions (freeze the subset while varying momentum, or keep optimizer state fixed while live-updating the subset) to attribute observed memory to the sampler vs.\ trajectory.

\subsection{RL replay: mature design space for prioritized, stateful sampling.}
Experience replay predates deep RL, framing memory as a buffer of transitions reused for sample efficiency \citep{Lin1992Replay}. In deep RL, replay is indispensable (e.g., DQN) and entails design choices that \emph{shape} learning dynamics: buffer size (an effective half-life), prioritization by TD-error with importance-sampling correction, and distributed pipelines that decouple acting from learning \citep{Mnih2015DQN,Schaul2016PER,Horgan2018ApeX}. Systematic studies show performance can swing with buffer size and prioritization schedules—clear, quantifiable manifestations of sampler memory and its lifetime \citep{ZhangSutton2017DeeperReplay,Fedus2020RevisitReplay}. Distributed agents (Ape-X, R2D2) surface further issues: parameter lag and representational drift between actors and the learner, showing that \emph{who} writes to the buffer and \emph{when} experiences are consumed are causal factors \citep{Horgan2018ApeX,Kapturowski2019R2D2}. The RL toolkit thus offers off-the-shelf levers (priority exponent, IS annealing, buffer turnover) and reporting patterns (return with CIs across seeds, ablations over buffer hyperparameters) that translate directly to supervised settings: a prioritized sampler is simply PER without TD-errors, and its “age” distribution is a measurable lifetime parameter.

\subsection{Federated optimization: cross-round memory via server/client state.}
Federated learning bakes memory into the protocol: local steps on non-IID client data, then global aggregation. FedAvg effectively introduces long \emph{round-scale} memory because models evolve locally before being mixed, and the server may accumulate its own momentum or adaptive moments \citep{mcmahan2017fedavg,reddi2021fedopt}. Client drift under heterogeneity motivated explicit control variates (SCAFFOLD), making cross-round corrections an inspectable state that reduces variance \citep{Karimireddy2020Mime}. FedProx tethers local objectives to the global iterate, an explicit \emph{path} constraint that stabilizes long-lived memory across rounds \citep{Li2020FedProx}. MIME shows how to mimic centralized momentum/Adam by sharing server-side statistics so that client updates track a target trajectory, again turning trajectory memory into an explicit, portable object \citep{Karimireddy2020Mime}. Methodologically, the federated literature is unusually disciplined about reporting across seeds, client participation, and heterogeneity settings, and often provides ablations that flip server momentum, control-variates, or local step counts one at a time—precisely the kind of causal diagnostics our section advocates.

\subsection{What transfers back.}
Across these areas, three themes recur. First, \emph{make memory visible}: record and expose the state that carries over (buffers, moments, control variates, selected subsets), so it can be reset or swapped. Second, \emph{treat lifetime as a parameter}: buffer size in RL, number of local steps in FL, and curriculum pacing in supervised learning are all tunable half-lives with measurable effects on endpoints. Third, \emph{evaluate with task-aligned, uncertainty-aware metrics}: continual learning’s backward/forward transfer, RL’s returns across seeds with CIs, and coreset generalization gaps all go beyond single-run accuracy. These are templates for the causal, function-space–aware measurements we argue are missing in standard training reports.

\section{Open Problems}
\label{sec:open-problems}

The case for \emph{training memory} is clear: optimizer moments, sampler policies, and the path a model takes through parameter space all leave measurable traces. What is missing is a way to study these traces that scales across modalities, survives replication, and speaks to both theory and practice. We articulate below a set of open problems that, if addressed, would turn scattered evidence into cumulative science. Each problem is framed to be solution-agnostic but empirically actionable.

\textbf{OP1. Causal attribution \emph{at scale} with standardized protocols.}
Small, carefully controlled studies can disentangle optimizer state from data order or augmentation phases, but we lack \emph{portable} protocols that make those attributions auditable across labs and modalities. The underspecification problem \citep{damour2022underspecification} shows that pipelines with identical headline accuracy can encode very different histories. Reproducibility programs have improved documentation practices \citep{pineau2021reproducibilityreport}, yet there is no community standard for memory-specific artifacts: e.g., \emph{order hashes} for each epoch, \emph{buffer-state logs} for momentum/Adam/EMA/SWA, and \emph{pre-registered perturbations} that toggle exactly one memory source at a time (cf.\ \S\ref{sec:desiderata}). An open problem is to define such protocol cards and checksums so that one group’s “AB$\neq$BA” claim is straightforward for another to replay and verify on different hardware, data sets, and architectures.

\textbf{OP2. Early diagnostics that \emph{predict across policies}, not just within them.}
We often observe early divergence in loss, calibration, or function-space distances, but when do these signals \emph{generalize} across momentum schedules, augmentation phases, or sampler policies? Learning-curve extrapolation and freeze–thaw approaches \citep{swersky2014freezethaw,domhan2015lce,klein2017lcp} forecast eventual accuracy, and data-centric early signals (forgetting events, EL2N) identify influential examples \citep{Toneva2019Forgetting,Paul2021EL2N}. Flatness-oriented training (SAM) \citep{foret2020sharpness}, large-batch generalization behavior \citep{keskar2017largebatch}, and double-descent phenomenology \citep{nakkiran2021deepdd} suggest that early geometry interacts strongly with later outcomes. The open question is to define early-window readouts—with uncertainty—that retain \emph{rank consistency and calibration} when we vary optimizer half-life, order policy, or augmentation schedule, and to formalize equivalence margins that support “no meaningful difference” conclusions when predictions do not transfer.

\textbf{OP3. Stateful sampling under label noise and distribution shift.}
Prioritizing by loss, gradient norm, or “learnability’’ can accelerate training \citep{katharopoulos2018importance,johnson2018training}, but under label noise or spurious correlations, such feedback can entrench errors. Robust learning-from-noise methods (e.g., MentorNet) \citep{jiang2020fantastic} and surveys \citep{Song2020NoisyLabels} document mitigations; coreset/subset selection gives principled alternatives (CRAIG, GLISTER, GRAD-MATCH) \citep{Mirzasoleiman2020CRAIG,Killamsetty2020GLISTER,Killamsetty2021GradMatch}. Still lacking is a \emph{unified diagnostic} that (i) estimates the bias/variance introduced by adaptive sampling with confidence intervals, (ii) stress-tests samplers under controlled class-conditional noise and realistic shifts (cf.\ WILDS) \citep{koh2021wilds}, and (iii) triggers \emph{on-the-fly} corrections (importance weights, annealed priorities) when early signals indicate drift toward mislabeled or out-of-distribution pockets.

\textbf{OP4. Deep-specific theory for non-commutativity and path dependence.}
We know stochastic updates need not commute in nonconvex objectives; order and path matter. There is geometric evidence—loss-landscape shape and mode connectivity \citep{li2018visualizing,garipov2018loss}—and stronger nonconvex SGD analyses \citep{mishchenko2020random}, but we lack a theory that \emph{quantifies} AB$\neq$BA effects \emph{in function space} under realistic deep-training ingredients: momentum/EMA half-lives, heavy augmentation, and stateful sampling. Open directions include linking measurable half-lives (momentum $\beta$, EMA decay, replay turnover) to bounds on non-commutativity; characterizing when different orderings remain linearly connected vs.\ diverge into isolated basins; and extending analyses from IID/reshuffle regimes to curricula and prioritized policies that evolve with the learner. Connections to implicit bias in deep models \citep{soudry2018implicit,gunasekar2018implicit,chizat2019lazy} and flatness-aware updates \citep{foret2020sharpness} remain to be made precise.

\textbf{OP5. Privacy and safety of explicit vs.\ implicit memory.}
Explicit memory (replay buffers, exemplar sets, server momentum snapshots) sharpens attribution but raises privacy and safety concerns: membership inference \citep{Shokri2017Membership}, training-data extraction \citep{Carlini2021Extraction}, and gradient inversion \citep{Zhu2019DLG}. Federated learning adds cross-round state and system-level visibility; secure aggregation and DP help but interact subtly with stateful optimization \citep{bonawitz2017secureagg,mcmahan2017fedavg,andrew2021dpclipping}. Open questions: how to design memory interventions that are \emph{privacy-aware by construction} (e.g., DP-safe buffer turnover, per-sample accounting for prioritized draws); how to quantify the trade-off between stateful gains and leakage risk under realistic threat models; and how to extend unlearning guarantees to both explicit buffers and path-dependent implicit memory.

\textbf{OP6. Robustness of memoryful training under shift.}
Distribution-shift benchmarks reveal gaps that accuracy alone hides \citep{koh2021wilds,ovadia2019can}. Do curricula, prioritized replay, or momentum half-life tuning improve or degrade calibration and OOD reliability? The open problem is a protocol that attributes OOD behavior to \emph{specific} memory sources, using probe-based function-space distances with uncertainty alongside standard metrics, and that reports \emph{per-subpopulation} effects so that “wins’’ are not averaged away.

\textbf{OP7. System-level memory in federated and distributed regimes.}
Client ordering, local step counts, and server optimizers imprint cross-round memory; control variates (SCAFFOLD) and proximal terms (FedProx) make that memory explicit \citep{Karimireddy2020Scaffold,Li2020FedProx,reddi2021fedopt}. We lack measurement tools that separate client sampler effects (non-IID order, augmentation) from server-side trajectory memory, and that expose compute–privacy–convergence trade-offs. Recent analyses (e.g., non-parametric views of FedAvg) \citep{su2023nonparametricfedavg} motivate diagnostics that are agnostic to parametric assumptions yet sensitive to round-scale history.

\textbf{OP8. Standardized uncertainty for training-dynamics claims.}
Finally, the community needs norms for uncertainty in \emph{training-memory} studies: multi-seed summaries; paired/resampled bootstrap over fixed probes; equivalence tests with declared margins for “no meaningful difference’’; and transparent pre-registration of which parameters will be varied (LR, $\beta$, batch size, order policy, state carry/reset) \citep{dror2018hitchhikers,pineau2021reproducibilityreport}. Without these, memory effects remain fragile to researcher degrees of freedom.

Progress will come from paired advances: protocol artifacts that make source isolation checkable at scale; early diagnostics validated across families of optimizers and samplers; robust, debiased stateful sampling; deep-theory links between half-lives and non-commutativity; and privacy-aware memory accounting. Each thread is orthogonal to any single diagnostic, but together they enable results that travel across data sets, labs, and modalities.

\section{Scope, Limitations, and Threats to Validity}
\label{sec:scope-limitations-threats}

\subsection{Scope \& non-claims.}
This article advances a \emph{measurement protocol} for attributing observed training effects to specific memory sources via minimal interventions and paired average treatment effect (ATE) estimands. It does \emph{not} propose a new learning algorithm, optimizer, loss, or architecture; it is \emph{not} a new benchmark; and it does \emph{not} introduce a novel statistical estimator beyond standard paired designs and nonparametric uncertainty quantification. We intentionally avoid prescriptive thresholds for what constitutes a “material” effect. Instead, we recommend reporting effect sizes with uncertainty (e.g., bootstrap confidence intervals over seeds/runs) and leaving materiality judgments to downstream users and domains. Our results are interventional within the studied training stacks and timescales; they do not claim causal transportability to unseen regimes, nor do they preclude unmeasured nuisance factors.

\subsection{Search Strategy and Survey Methodology}
\label{sec:methodology}

This work is an \emph{expert-guided scoping survey} with a structured search protocol. The initial corpus was assembled from field knowledge and canonical seed papers, then expanded via database queries and forward/backward passes. We queried multiple scholarly indexes to mitigate source bias:

\begin{itemize}
  \item \textbf{Computer science indices:} ACM Digital Library, IEEE Xplore, DBLP, arXiv (cs.LG, cs.CV, cs.AI), and Google Scholar.
  \item \textbf{General indices:} Scopus and Web of Science (for cross-disciplinary coverage).
\end{itemize}
\textbf{Time window:} January 1, 2010 through September 23, 2025 (inclusive). We include pre-2010 optimization classics when directly relevant (e.g., heavy-ball momentum, Nesterov acceleration).

We grouped queries by our taxonomy’s \emph{source} axis; each block was executed with field restrictions (Title/Abstract/Keywords when supported) and then as a full-text fallback. Representative queries follow (parentheses indicate OR-lists; wildcards as supported by the engine):

\paragraph{S1: Optimizer/trajectory state.} \begin{verbatim}
("momentum" OR "Nesterov" OR "heavy-ball" OR "Polyak")
AND ("deep neural" OR "neural network" OR "deep learning")
AND (training OR optimization)

("Adam" OR "AdamW" OR "adaptive moment" OR "decoupled weight decay")
AND ("deep" OR "neural") AND (generalization OR calibration OR "test error")

("exponential moving average" OR EMA OR "stochastic weight averaging" OR SWA)
AND (neural OR deep) AND (generalization OR calibration)

("sharpness-aware" OR SAM) AND (generalization OR robustness)

("K-FAC" OR "Kronecker-factored" OR Shampoo OR "second-order" OR precondition*)
AND (neural OR deep)
\end{verbatim}

\paragraph{S2: Sampler/data-order state.}
\begin{verbatim}
("random reshuffling" OR "without replacement" OR "with replacement"
 OR "data order" OR "minibatch order")
AND (stochastic gradient* OR SGD) AND (deep OR neural)

(curriculum OR pacing OR "self-paced learning" OR "staged augmentation")
AND ("deep learning" OR "neural network")

("importance sampling" OR "prioritized sampling" OR "priority sampling")
AND (deep OR neural OR SGD)

(replay OR "experience replay" OR coreset OR "memory buffer")
AND (continual OR incremental OR "class-incremental" OR deep)
\end{verbatim}

\paragraph{S3: Parameter-path dependence.}
\begin{verbatim}
("mode connectivity" OR "loss landscape" OR "flat minima" OR "flatness"
 OR "path dependence" OR noncommutativ* OR "AB!=BA")
AND ("deep neural" OR "neural network")
\end{verbatim}

\paragraph{S4: Architectural/external memory.}
\begin{verbatim}
("memory bank" OR queue OR "feature bank" OR "momentum encoder" OR MoCo
 OR "instance discrimination" OR Hebbian OR "plasticity trace")
AND ("self-supervised" OR contrastive OR "deep learning")
\end{verbatim}

\paragraph{S5: Meta-state.}
\begin{verbatim}
("mean teacher" OR "teacher EMA" OR "temporal ensembling" OR "lookahead optimizer"
 OR "learned optimizer" OR "meta-optimizer")
AND ("deep learning" OR "neural network")
\end{verbatim}

\paragraph{Cross-cutting measurement terms (combined with blocks above).}
\begin{verbatim}
AND (generalization OR calibration OR "representation similarity" OR SVCCA OR CKA
 OR "function space" OR "prediction distribution" OR "causal" OR "intervention")
\end{verbatim}

\subsection{Eligibility criteria}
We followed a two-stage screen with de-duplication:
\begin{enumerate}
  \item \textbf{Title/abstract screen} by one reviewer; borderline items retained.
  \item \textbf{Full-text assessment} for relevance to at least one source axis (S1–S5) and at least one lifetime (step/epoch/phase/task).
\end{enumerate}
\textbf{De-duplication} used DOI and normalized titles across databases. \textbf{Forward/Backward passes} added forward citations and backward references from included “seed” papers, capped when no new items passed the following (Inclusion)–(Exclusion) criteria for two consecutive waves.

\textbf{Inclusion:}
\begin{enumerate}
  \item Peer-reviewed conference/journal papers or widely adopted preprints (clear community uptake) in ML/AI/vision/NLP.
  \item The work \emph{introduces, analyzes, or empirically studies} mechanisms that carry state across updates/epochs/phases (optimizers, samplers, path/geometry, external memory, meta-state) \emph{or} measures their effects (e.g., order dependence, averaging, representation drift).
  \item For privacy/robustness (e.g., membership inference, data extraction), included when the phenomenon is tied to training history or state retention in the model.
\end{enumerate}
\textbf{Exclusion:}
\begin{enumerate}
  \item Theses, blogs, patents, and purely tutorial/introductory pieces without original empirical/theoretical contribution.
  \item Works on reinforcement learning where replay is used solely for off-policy control without claims about training-memory mechanisms relevant to supervised/self-supervised training.
  \item Papers whose only use of “memory” is architectural (e.g., LSTMs/Transformers) without updates to external state during training beyond standard weights/optimizer (unless they explicitly act as training-time memory banks).
\end{enumerate}

\subsection{Data extraction and coding schema}
For each included paper we recorded:
\begin{itemize}
  \item \textbf{Bib/venue/year} and \textbf{area}.
  \item \textbf{Taxonomy labels:} source (S1–S5), lifetime (step/epoch/phase/task), visibility (explicit/implicit).
  \item \textbf{Setting:} supervised, self-supervised, continual, federated; dataset/task family.
  \item \textbf{Mechanism studied:} e.g., momentum/EMA/SWA; RR vs. replacement; curriculum; prioritization; replay; mode connectivity; queues/banks; teacher EMA; learned optimizer.
  \item \textbf{Measurement:} metrics (accuracy, loss, calibration, SVCCA/CKA, function-space distances), whether \emph{causal interventions} were performed (carry vs. reset, order swaps), and whether uncertainty was reported (CIs, tests).
  \item \textbf{Findings (qualitative)}.
\end{itemize}

\subsection{Surveyal Limitations}
Terminology around “memory” varies (e.g., implicit bias vs. flatness vs. order effects), which risks false negatives under strict keywording; we mitigated this with snowballing and cross-area seeds. We prioritize works that either analyze state carry-over or \emph{measure} its impact; adjacent but orthogonal topics (e.g., architecture-only memory without training-time state changes) are not exhaustively covered.

\subsection{Threats to validity}
We phrase threats and safeguards in terms of abstract components that any implementation provides: the \emph{execution substrate} (numerical kernels and schedulers), \emph{pseudo-random number generators} (PRNGs), the \emph{data-order mechanism} (how example indices are sequenced), \emph{evaluation probes} (fixed inputs for function-space readouts), and \emph{normalization layers with running state}. These abstractions keep the guidance independent of programming languages, libraries, and hardware (see \tableautorefname~\ref{tab:threats-validity}).
\begin{table}[t]
\centering
\scriptsize
\setlength{\tabcolsep}{3pt}
\renewcommand{\arraystretch}{0.96}
\begin{tabularx}{\linewidth}{l X X X}
\toprule
\textbf{Threat} & \textbf{Failure Mode} & \textbf{Impact} & \textbf{Mitigation} \\
\midrule
Numerical nondeterminism &
Non-reproducible kernel paths or small numeric drift across environments. &
Higher within-branch variance; spurious cross-branch deltas; weaker paired ATEs. &
Record a substrate manifest (versions/digests); prefer deterministic modes; disable kernel autotuning; use named PRNG streams with fixed seeds; note which parts are best-effort. \\
\addlinespace[1pt]
Order drift (data-order mech.) &
Parallel loading/partitioning/resume silently changes example-index sequences across epochs/branches. &
Unintended S2/S3 perturbations masquerade as the intended intervention. &
Derive per-epoch permutations from explicit (seed, epoch) contracts; log short digests per worker/partition to verify equality; freeze worker seeding/resume semantics; log sampler config verbatim. \\
\addlinespace[1pt]
Probe cross-talk &
Probes (held-out batches, cached activations, retrieval keys, feature buffers) shared via caches/global state. &
Contaminated measurements; shrunk or reversed deltas. &
Materialize branch-local probe copies; isolate caches and PRNGs for eval; include probe content digests; treat retrieval/teacher/external-memory buffers as branch-scoped. \\
\addlinespace[1pt]
Normalization miscalibration &
Interventions change features while normalization layers with running state retain stale stats. &
Apparent gains/losses due to stale normalization, not the intended source. &
Re-estimate running stats on a held-out calibration pass \emph{per branch} before final eval, or explicitly freeze and state policy; for EMA/SWA, declare calibration data and passes. \\
\bottomrule
\end{tabularx}
\caption{Threats framed as components.}
\label{tab:threats-validity}
\end{table}

\section{Conclusion}
Training memory is a feature, not a bug, of modern pipelines. Momentum and adaptive moments, weight and teacher averaging, data-order policies and staged augmentations, architectural side memory (queues, replay), normalization statistics, and nonconvex path dependence all carry history across updates. Our synthesis argues that these sources should be treated as \emph{distinct}, with characteristic lifetimes and with varying degrees of visibility and resetability. The practical consequence is methodological: claims about training dynamics and generalization are persuasive only when they (i) perturb one source at a time, (ii) verify controls, and (iii) read out effects in function space with uncertainty.

We offered four structures to make this routine. The \emph{source–lifetime–visibility} taxonomy clarifies where memory comes from and how long it lasts. The consolidation of theory and empirical patterns explains why order and optimizer state can independently move both trajectory and endpoint. The \emph{causal estimands} and \emph{perturbation primitives} turn attribution into seed-paired experiments with portable interventions and principled uncertainty (bootstrap CIs and equivalence testing). Finally, the reporting checklist enumerate what must be fixed, logged, and shared—order hashes, optimizer/EMA/BN snapshots, queue state, RNG stream contracts, and configuration artifacts—to make results auditable and replayable.

We see three near-term payoffs. First, sharper \emph{attribution}: studies can separate optimizer vs.\ sampler vs.\ path effects rather than bundling them into a single headline gain. Second, better \emph{predictivity}: early, function-space signals can be tested for rank stability \emph{across} policies, not only within them. Third, improved \emph{transfer and safety}: memory-aware protocols clarify how phase boundaries, replay, and server-side adaptivity affect calibration, robustness, and privacy.

Limitations remain. Our recommendations do not remove all nondeterminism, and they do not resolve deep-theory gaps in noncommutativity or the privacy risks of explicit memory (replay, server accumulators). They also ask for modest but real engineering discipline. Still, the bar is attainable on commodity budgets: small, well-instrumented testbeds with seed pairing and audit logs provide outsized clarity.

The path forward is communal. If authors routinely release audit artifacts alongside code, adopt single-source perturbations with seed-paired ATEs, and report function-space deltas with uncertainty, “training memory” will transition from folklore to cumulative science. We hope this survey accelerates that transition by supplying the vocabulary, contrasts, and minimal tooling to make memory \emph{measurable} and \emph{auditable}.

\bibliographystyle{unsrt}  
\bibliography{references}

\end{document}